%% file: main.tex
\documentclass[runningheads]{llncs}

\usepackage{eccv}

\newcommand{\mb}[1]{\mathbf{#1}}

\usepackage{eccvabbrv}
\usepackage{graphicx}
\usepackage{booktabs}
\usepackage[accsupp]{axessibility}
\usepackage{hyperref}
\usepackage{orcidlink}
\usepackage[table]{xcolor}
\usepackage{comment}

\usepackage{multirow}

\begin{document}

\title{LSRM: High-Fidelity Object-Centric Reconstruction via Scaled Context Windows} 

\titlerunning{High-Fidelity Object-Centric Reconstruction via Scaled Context Windows}

\author{Zhengqin Li \and
Cheng Zhang \and
Jakob Engel \and
Zhao Dong
}

\authorrunning{Z.~Li et al.}
\institute{Meta Reality Labs Research}

\maketitle

\begin{abstract}
We introduce the \textbf{L}arge \textbf{S}parse \textbf{R}econstruction \textbf{M}odel to study how scaling transformer context windows affects feed-forward 3D reconstruction. Although recent object-centric feed-forward methods produce robust, high-quality reconstructions, they still lag behind dense-view optimization in recovering fine-grained texture and appearance. We show that expanding the context window---by substantially increasing the number of active object and image tokens---narrows this gap and enables high-fidelity 3D object reconstruction and inverse rendering. To scale effectively, we adapt native sparse attention \cite{yuan2025native} for 3D reconstruction with three key contributions: (1) an efficient coarse-to-fine pipeline that focuses computation on informative regions by predicting sparse high-resolution residuals; (2) a 3D-aware spatial routing mechanism that establishes accurate 2D-3D correspondences using explicit geometric distances rather than standard attention scores; and (3) a custom block-aware sequence-parallel strategy with an All-gather-KV protocol to balance dynamic, sparse workloads across GPUs. As a result, LSRM handles $\mathbf{20\times}$ more object tokens and >$\mathbf{2\times}$ more image tokens than prior state-of-the-art (SOTA) methods. Extensive evaluations on standard novel-view synthesis benchmarks show substantial gains over the current SOTA, yielding >$\mathbf{2.4}$~dB higher PSNR and >$\mathbf{40\%}$ lower LPIPS. Furthermore, when extending LSRM to inverse rendering, qualitative and quantitative evaluations on widely used benchmarks demonstrate consistent improvements in texture and geometry details, achieving an LPIPS that matches or exceeds that of SOTA dense-view optimization methods. Code and model weights are available on our \href{https://lzqsd.github.io/LSRM.github.io/}{project page}.

\end{abstract}

\input{section/intro}

\input{section/related}
\input{section/method}
\input{section/experiments}
\input{section/conclusions}

\clearpage
\appendix
\section*{Supplementary Material}

This appendix provides additional implementation details and extended evaluations:
\begin{itemize}
    \item \textbf{Native Sparse Attention:} We describe the attention branches used by NSA and discuss the compressed KV layout required for efficient Triton kernels.
    \item \textbf{Training details and losses:} We report stage-wise hyperparameters and the loss functions used for novel view synthesis and inverse rendering.
    \item \textbf{Additional evaluation results:} We provide depth and normal comparisons, robustness analysis for Stage 1 geometric errors, and comparisons with 3D Gaussian optimization.
\end{itemize}

\input{section/nsa}
\input{section/details}
\input{section/evaluation}

\bibliographystyle{splncs04}
\bibliography{main}
\end{document}

%% file: section/intro.tex
\section{Introduction}
\label{sec:intro}

\begin{figure}[t]
\centering
\includegraphics[width=\linewidth]{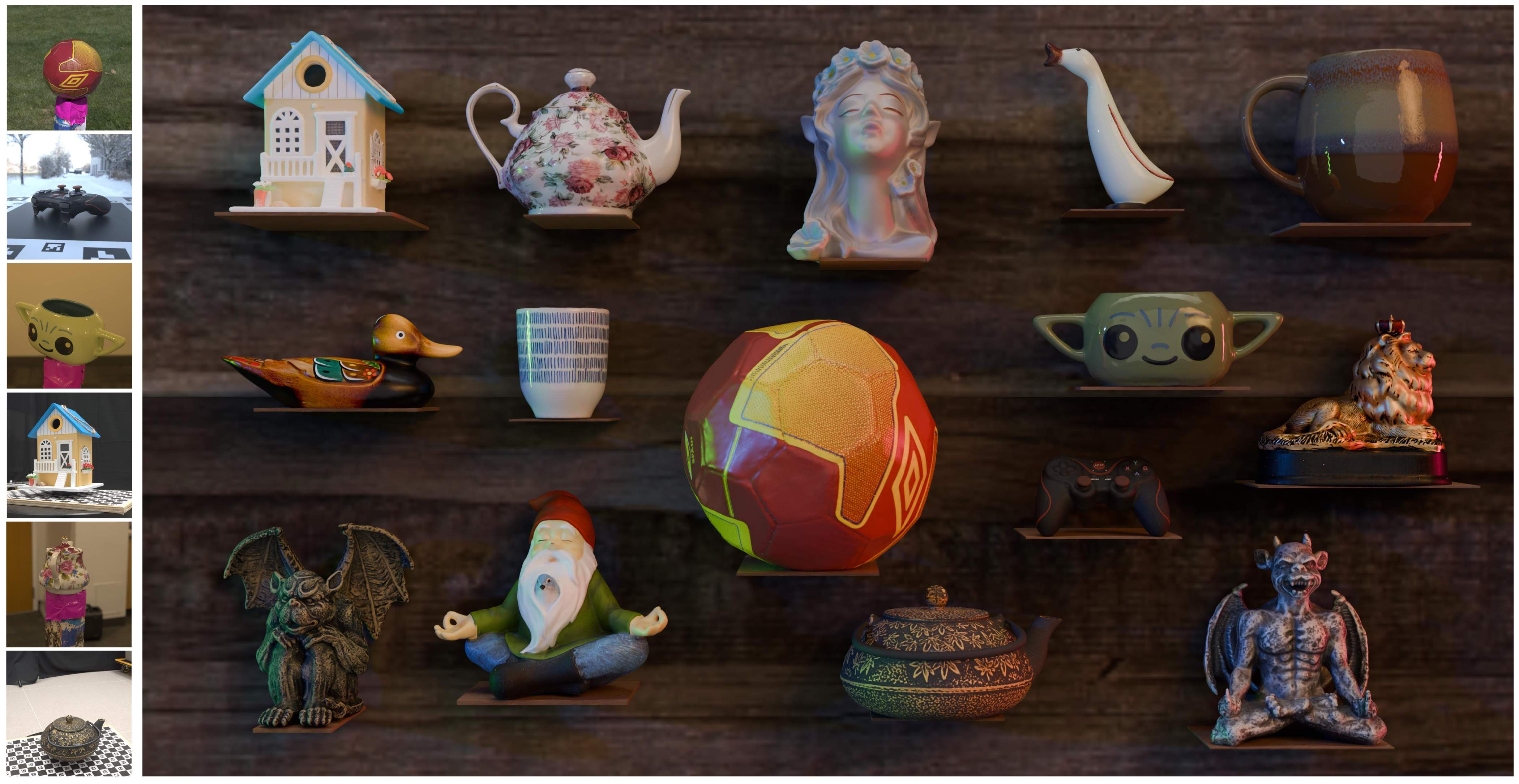}
\caption{\textbf{High-fidelity 3D reconstruction.} Given 6--18 images (\textbf{left}), LSRM adapts Native Sparse Attention (NSA) to generate explicit meshes and textures in a single feed-forward pass with minimal compute overhead. As an extension, LSRM can also predict BRDF maps for inverse rendering (\textbf{right}). Zoom in for details.}
\label{fig:teaser}
\end{figure}

Recent years have witnessed rapid progress in the application of 3D foundation models---typically built on large-scale transformer architectures \cite{vaswani2017attention}---to tackle 3D tasks previously considered intractable. These tasks include joint estimation of geometry and camera parameters \cite{wang2024dust3r,leroy2024grounding,lin2025depth,wang2025vggt}, dynamic scene reconstruction \cite{lin2025dgs,xu20254dgt,ma20254d,liang2024feed,zhang2025efficiently}, and sparse-view reconstruction \cite{hong2023lrm,wei2024meshlrm,zhang2025gs,jin2024lvsm,zhang2024relitlrm} and inverse rendering \cite{li2025lirm,zarzar2025twinner}. In object-centric reconstruction and inverse rendering in particular, feed-forward models cut reconstruction time by orders of magnitude compared with optimization-based pipelines \cite{sun2023neural,zhang2022iron,zhang2021physg,munkberg2022extracting}, while achieving competitive view synthesis and relighting quality, and often exhibiting stronger geometric robustness on challenging specular objects \cite{li2025lirm}. Yet a key limitation persists: reconstructed textures frequently miss fine-grained detail, leading to blurry text, smeared logos, and distorted facial features. Recent attempts to address this---by swapping the underlying 3D representation \cite{zhang2025gs,jin2024lvsm} or adding post-hoc texture refinement \cite{siddiqui2024meta}---improve results but still do not match the fidelity of dense-view optimization. Narrowing this gap is essential for deploying feed-forward 3D reconstruction at the quality and reliability required for practical, industrial-scale 3D digital twin creation.

We hypothesize that this gap can be narrowed by expanding the context window---specifically, by increasing the number of active object and image tokens. However, na\"ively upscaling the resolution of 3D object or 2D image representations rapidly becomes impractical, incurring quadratic (and in volumetric settings, cubic) growth in GPU memory and compute. To overcome this bottleneck, we adapt native sparse attention (NSA) \cite{yuan2025native}, a hardware-friendly algorithm that restricts each token's attention to a sparse set of token blocks, enabling efficient training and inference at substantially larger context sizes. Building on NSA, we develop an efficient coarse-to-fine training pipeline. In the first stage, a dense reconstruction transformer identifies the most informative volume tokens; in parallel, we prune 2D image tokens by retaining only those covering foreground pixels.  This coarse model provides a strong initialization for both the network weights and the geometric structure. In the second stage, the sparse reconstruction transformer refines the reconstruction by predicting high-resolution sparse volume residuals, an approach we find considerably more effective than directly regressing a high-resolution volume from scratch.

While enlarging the context window with NSA already improves baseline quality, we introduce two critical architectural adaptations to further maximize reconstruction fidelity and hardware efficiency. First, we observe that the default NSA block selection \cite{yuan2025native}---which relies purely on attention scores from compressed tokens---often fails to capture reliable correspondences between voxel and image tokens, especially in early transformer layers. To remedy this, we propose a 3D-aware spatial routing strategy that explicitly uses the geometric distances established in Stage 1, alongside known camera parameters, to retrieve relevant blocks. This 3D-aware routing markedly enhances texture sharpness, rendering previously blurred fine text legible. Second, the dynamic sparse layout naturally induces severe token imbalance across GPUs. To address this, we design a custom block-aware sequence parallelism scheme. We shard tokens by their spatial 2D and 3D blocks, ensuring tokens within the same block remain localized on the same GPU to eliminate cross-device communication during local attention. Furthermore, we leverage the high KV compression of Grouped-Query Attention (GQA) \cite{ainslie2023gqa} to implement an All-gather-KV protocol. This efficiently broadcasts a lightweight KV cache across the node, providing full global context with minimal communication overhead. Together, these designs enable LSRM to process $\mathbf{20\times}$ more object tokens and >$\mathbf{2\times}$ more image tokens than the prior SOTA \cite{li2025lirm}, without a proportional increase in computational resources. Comprehensive experiments on standard object-centric 3D reconstruction benchmarks demonstrate substantial gains, yielding >$\mathbf{2.4}$~dB higher PSNR and a >$\mathbf{40\%}$ reduction in LPIPS on the GSO dataset \cite{downs2022google}. We further extend LSRM to the task of inverse rendering by fine-tuning the model to predict material reflectance from images captured under natural illumination. Both qualitative and quantitative experiments show significant improvements in textural and geometric detail. Specifically on the widely used StanfordORB dataset \cite{kuang2024stanford}, LSRM sets a new standard for feed-forward inverse rendering methods, achieving an LPIPS that matches the SOTA dense-view optimization method \cite{sun2023neural}.

%% file: section/related.tex
\section{Related Work}

\paragraph{Object-Centric Feed-Forward Reconstruction} 3D foundation models trained on large-scale datasets \cite{deitke2023objaverse} have driven major progress in object-centric reconstruction. Pioneering models like LRM \cite{hong2023lrm} achieved impressive single-image geometry recovery, but they often struggle to reproduce high-fidelity textures. Subsequent works have sought to improve appearance quality through architectural refinements \cite{wei2024meshlrm}, alternative 3D representations \cite{zhang2025gs,xu2024grm,ziwen2025long,jin2024lvsm,chen2024lara}, progressive updates \cite{li2025lirm}, and post hoc texture refinement \cite{siddiqui2024meta}. Despite these advances, a sizable performance gap compared to dense-view, optimization-based pipelines persists. A recent approach \cite{zhang2025test} narrows this gap via test-time training to effectively enlarge the context window; however, it lacks an explicit 3D representation and may suffer from view synthesis degradation at unfamiliar camera angles. In contrast, LSRM leverages scaled context windows to explicitly reconstruct high-fidelity 3D geometry and textures in a single forward pass. Furthermore, by extending LSRM to predict BRDF material maps, we enable seamless downstream editing and realistic relighting within standard graphics pipelines.

\paragraph{Inverse Rendering} Inverse rendering aims to decompose images into their intrinsic factors---geometry, materials, and illumination---so that scenes or objects can be edited and re-rendered realistically. Classical measurement-based approaches \cite{debevec2000acquiring,bi2020deep,goldman2009shape} rely on specialized capture rigs and controlled environments to densely sample view and lighting directions. Recent optimization-based methods \cite{zhang2021nerfactor,zhang2021physg,zhang2022iron,boss2021nerd,boss2021neural,boss2022samurai,engelhardt2024shinobi,zhang2022modeling,jin2023tensoir,munkberg2022extracting,hasselgren2022shape,sun2023neural,liang2024gs,jiang2024gaussianshader,zhang2023nemf,yang2023sire,shi2023gir} can often work under natural illumination, but they usually require dense multi-view inputs and remain fragile in ill-posed conditions such as saturated specularities, cast shadows, and complex interreflections. Many learning-based techniques focus on predicting per-pixel BRDF maps \cite{careaga2023intrinsic,meka2018lime,li2018cgintrinsics,li2018learning,li2018materials,li2017modeling,deschaintre2018single,li2022physically,zeng2024rgb} without producing a complete 3D representation. Recently, propelled by advances in object-centric 3D reconstruction models, pioneering works \cite{li2025lirm,zarzar2025twinner} have shown that feed-forward models can output fully relightable 3D objects with explicit geometry and materials. LSRM builds on this direction and further improves reconstruction and appearance fidelity by enabling substantially longer context windows.

\paragraph{Sparsity in 3D Generation and Reconstruction} Many prior approaches handle the inherent sparsity of 3D surfaces indirectly, e.g., by compressing geometry into compact representations \cite{wu2024direct3d,lan2025ln3diff++,gupta20233dgen,zhang20233dshape2vecset,zhang2024clay,zhao2023michelangelo,li2024craftsman3d,li2025step1x,li2025triposg,hunyuan3d2025hunyuan3d} or adopting localized point-based formats such as 3D Gaussians \cite{chen20253dtopia,lan2025gaussiananything,vahdat2022lion,yang2024atlas}. More recently, a line of 3D generative work has modeled sparsity explicitly through structured latent representations---such as sparse voxel grids---to scale volumetric resolution substantially \cite{xiang2025structured,xiang2025native,li2025sparc3d,wu2025direct3d}. In these frameworks, sparse volumes are either encoded by sparse 3D VAEs into dense latent codes \cite{xiang2025structured,xiang2025native,li2025sparc3d,chen2025sam} or, as in Direct3D-S2 \cite{wu2025direct3d}, downsampled to a lower-resolution sparse volume that can be generated with an NSA-style transformer operating over 3D token blocks. While these methods demonstrate efficiency in processing high-resolution volumes, their focus remains restricted to synthesizing intricate geometric details, often neglecting texture. Even when texture generation is supported \cite{xiang2025native,chen2025sam}, the resulting texture fidelity remains limited even under clean synthetic inputs.
Such single-view generative models target plausible 3D appearance synthesis rather than high-fidelity deterministic reconstruction from posed multi-view inputs, making them complementary to our setting.
This highlights that high-fidelity textured object reconstruction is a distinct and challenging problem, motivating the specialized architecture design by LSRM.

%% file: section/method.tex
\section{Method}
\label{sec:method}

This section details the architecture of LSRM. We first formulate the Stage 1 dense reconstruction transformer, specifying the model's inputs, outputs, and notation. We then describe our Stage 2 sparse reconstruction transformer, which is initialized from the dense model and adapts NSA to significantly expand the context window. Finally, we detail two key improvements that maximize reconstruction fidelity and hardware efficiency. Additional background on NSA is provided in Appendix~\ref{sec:app_nsa}.

\begin{figure}[t]
\centering
\includegraphics[width=\linewidth]{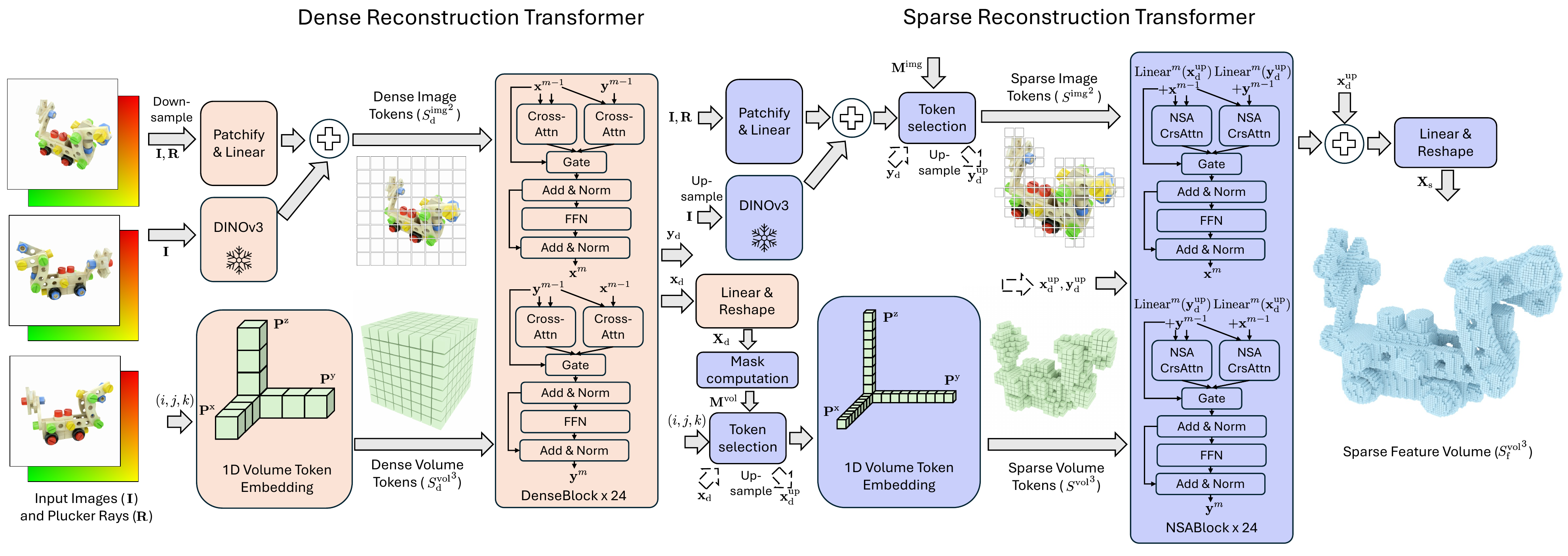}
\caption{\textbf{Network architecture of LSRM.} Our method employs a two-stage coarse-to-fine pipeline. In Stage 1, a Dense Reconstruction Transformer generates a coarse, low-resolution volume. Stage 2 utilizes this coarse volume to initialize the active sparse volume tokens and predicts high-resolution sparse residuals, constructing the final 3D representation.}
\label{fig:network}
\end{figure}

\subsection{Stage 1: Coarse Dense Reconstruction for Initialization}
We first train a dense reconstruction transformer for lower resolution reconstruction; its weights and intermediate predictions are then used to initialize the Stage 2 sparse reconstruction transformer. 

\paragraph{Image and Volume Tokenization} The inputs consist of a sparse set of $M$ posed images ($M\in [12, 18]$ similar to \cite{li2025lirm}), with camera parameters explicitly encoded as Pl\"ucker rays. Let $\{\mb{I}_m\}_{m=1}^M$ denote the non-overlapping $8 \times 8$ patches extracted from the $256 \times 256$ input views, yielding a spatial token map of resolution $S^\text{img}_\text{d} = 32$ per view. Let $\{\mb{R}_m\}_{m=1}^M$ denote their corresponding Pl\"ucker rays. We compute the combined multi-view image tokens, $\mb{y}$, by fusing a linear projection of the patches and rays with features extracted from a frozen DINOv3 \cite{simeoni2025dinov3} encoder:
\begin{equation}
    \mb{y} = \left\{ \text{Linear}(\mb{I}_m, \mb{R}_m) + \text{Linear}(\text{DINOv3}(\text{Upsample}(\mb{I}_m))) \right\}_{m=1}^{M}
\end{equation}
Because DINOv3 natively operates on $16 \times 16$ patches, the input images are upsampled by 2 prior to encoding. Incorporating DINOv3 features significantly accelerates training convergence and improves the model's generalizability. 

The volume tokens $\mb{x}$ are initialized from learned 3D positional embeddings. To avoid wasting GPU memory when later scaling to high-resolution grids in the Stage 2 sparse transformer, we factorize the embeddings along the three spatial axes. Instead of learning a dense parameter volume, we learn three independent 1D embeddings: $\mb{P}^\text{x} \in \mathbb{R}^{S^\text{vol}_\text{d} \times d}$, $\mb{P}^\text{y} \in \mathbb{R}^{S^\text{vol}_{\text{d}} \times d}$, and $\mb{P}^\text{z} \in \mathbb{R}^{S^\text{vol}_\text{d} \times d}$, where $d=1024$, $S^\text{vol}_\text{d}=16$ are the embedding dimension and the dense token volume's resolution. For a token at spatial coordinate $(i, j, k)$, its embedding $\mb{p}_{i,j,k}$ is simply the element-wise sum:
\begin{equation}
    \mb{p}_{i,j,k} = \mb{P}^\text{x}_i + \mb{P}^\text{y}_j + \mb{P}^\text{z}_k
\end{equation}
The resulting initialized volume tokens, $\mb{x} = \{\mb{p}_{i,j,k}\}$, are then concatenated with image tokens $\mb{y}$ and processed by the dense transformer blocks.

\paragraph{Dense Reconstruction Transformer} Following prior works \cite{wei2024meshlrm, li2025lirm}, our dense transformer consists of 24 blocks with hidden dimension 1024. Unlike prior dense models, we use grouped-query attention with 32 query heads and only 2 key/value heads, providing the high KV-head compression required by the subsequent sparse transformer and exploited by our sequence-parallel training (Sec. \ref{sec:seqp}). Additionally, rather than standard monolithic self-attention, we use a gated mixture of cross-attention updates to evolve token representations. This decoupled design provides flexible, fine-grained control over attention branches in Stage 2—for example, allowing independent tuning of the number of selected KV-blocks ($L$) for image and volume tokens. Let $\mb{x}$ denote volume tokens and $\mb{y}$ denote image tokens. The attention module within one DenseBlock is:
\begin{align}
    \mb{o}^{\mb{x}} &= \mb{w}^{\text{x-self}} \text{CrossAttn}(\mb{x}, \mb{x}) + \mb{w}^{\text{x-cross}} \text{CrossAttn}(\mb{x}, \mb{y}) \label{eq:crossattn_1} \\
    \mb{o}^{\mb{y}} &= \mb{w}^{\text{y-self}} \text{CrossAttn}(\mb{y}, \mb{y}) + \mb{w}^{\text{y-cross}} \text{CrossAttn}(\mb{y}, \mb{x}) \label{eq:crossattn_2} \\
    \mb{w}^{\text{x-self}}, \mb{w}^{\text{x-cross}} &= \text{Sigmoid}(\text{Linear}_{\text{gate}}^{\mb{x}}(\mb{x})) \\
    \mb{w}^{\text{y-self}}, \mb{w}^{\text{y-cross}} &= \text{Sigmoid}(\text{Linear}_{\text{gate}}^{\mb{y}}(\mb{y}))
\end{align}
We denote by $\mb{x}_\text{d}$ and $\mb{y}_\text{d}$ the volume and image token outputs from the final DenseBlock. We decode the volume tokens $\mb{x}_\text{d}$ into a dense feature volume via a linear projection. This layer upsamples spatial resolution by $4\times$, producing a grid of size $S^\text{vol}_\text{df} = 64$ while reducing feature dimension to $d_\text{f} = 32$:
\begin{equation}
    \mb{X}_\text{d} = \text{Linear}(\mb{x}_\text{d}), \quad \mb{X}_\text{d} \in \mathbb{R}^{{S^\text{vol}_\text{df}}^3 \times d_\text{f}} \label{eq:upsample}
\end{equation}

This dense feature volume, $\mb{X}_\text{d}$, can be utilized for either novel view synthesis or explicit textured mesh extraction via lightweight MLP decoders. With a slight abuse of notation, let $\mb{p} \in \mathbb{R}^3$ be a continuous 3D point sampled within the volume. We compute its corresponding properties as follows:
\begin{align}
    \mb{f} &= \text{Trilinear}(\mb{X}_\text{d}; \mb{p}) \\
    \mb{z} &= \text{Sigmoid}(\text{MLP}_{\mb{z}}(\mb{f})), \quad \mb{z} \in \{\mb{a}, \mb{r}, \mb{m}\} \text{ or } \mb{z} = \mb{c} \label{eq:mlpdecode_1} \\
    s &= \text{MLP}_{s}(\mb{f}) + s_\text{bias}(\mb{p}) \label{eq:mlpdecode_2}
\end{align}
where $\mb{f}$ is the trilinearly interpolated feature vector, $s$ is the predicted SDF, and $s_\text{bias}$ is an effective offset adapted from \cite{li2025lirm}. $\mb{c}$ denotes the emitted color for novel view synthesis. Alternatively $\mb{a}$, $\mb{r}$, and $\mb{m}$ represent the albedo, roughness, and metallic material properties for inverse rendering. For rendering, we adopt the VolSDF \cite{yariv2021volume} formulation due to its simplicity and effectiveness.

\subsection{Stage 2: Sparse Reconstruction with Long Context Windows}

For high-fidelity reconstruction in Stage 2, we substantially increase the resolutions of both the input images (e.g., $S^\text{img} = 3S^\text{img}_\text{d}=96$) and the volume grid (e.g., $S^\text{vol} = 6S^\text{vol}_\text{d}=96$). To efficiently handle the resulting expanded token sequence, we first describe how we extract active, spatially-sparse token subsets, and then present the spatial block partitioning required by our NSA mechanism, followed by the architecture of the sparse reconstruction transformer.

\paragraph{Informative Token Selection} 
We build a spatially-sparse token representation by retaining only the most informative image and volume tokens. For image tokens, we discard the background, preserving only patches that contain foreground pixels based on the foreground mask $\mb{M}^\text{img}$. For the volume tokens, we leverage the geometric prior from the Stage 1 dense transformer to compute a binary mask $\mb{M}^\text{vol}$. This mask identifies and retains only the voxels near the object surface, as these primarily determine the final appearance. Concretely, within a voxel at spatial coordinate $(i, j, k)$, we uniformly sample a grid of $T=4^3=64$ points $\mathcal{P}_{i,j,k} = \{\mb{p}_t\}_{t=1}^T$. We evaluate the SDF value $s_t$ at each point using the frozen Stage 1 dense feature volume $\mb{X}_\text{d}$ and set the voxel mask $\mb{M}^\text{vol}_{i,j,k}$ based on a distance threshold $\tau$:
\begin{align}
    s_t &= \text{MLP}_s(\text{Trilinear}(\mb{X}_\text{d}; \mb{p}_t)) \nonumber \\ 
    \mb{M}_{i,j,k}^\text{vol} &= \mathbb{I} \Big( \min_{\mb{p}_t \in \mathcal{P}_{i,j,k}} |s_t| \le \tau \quad\text{or}\quad \big( \min_{\mb{p}_t \in \mathcal{P}_{i,j,k}} s_t \big) \cdot \big( \max_{\mb{p}_t \in \mathcal{P}_{i,j,k}} s_t \big) \le 0 \Big) \label{eq:surface_voxel_mask}
\end{align}
where $\mathbb{I}(\cdot)$ denotes the indicator function. This rule ensures that only voxels close to the underlying surface, or directly intersected by it, are instantiated as tokens for high-resolution processing.

\paragraph{Block Partitioning and Compression}
A na\"\i ve 1D sequence partitioning of tokens destroys spatial locality, which is detrimental to fine-grained 3D reconstruction. Therefore, we adopt a spatial block partitioning strategy similar to Direct3D-S2 \cite{wu2025direct3d}. We group active image and volume tokens based on their original 2D and 3D coordinates. Specifically, space is divided into discrete blocks of size $8 \times 8$ for images and $8 \times 8 \times 8$ for the volume, reducing the block-level resolutions to $S_\text{b}^\text{img} = S^\text{img} / 8$ and $S_\text{b}^\text{vol} = S^\text{vol} / 8$. Because uninformative tokens are removed, each spatial block contains a variable number of active tokens. To compute the block-level compressed keys ($\mb{k}^\text{cmp}$) and values ($\mb{v}^\text{cmp}$) required by NSA, we first pass the individual token keys $\mb{k}_t$ and values $\mb{v}_t$ within a block through a residual module ($\text{ResBlock}$), and then average the resulting features in-block ($\text{AvgPool}$):
\begin{equation}
    \mb{k}^\text{cmp} = \text{AvgPool}(\text{ResBlock}(\mb{k}_t)), \quad \mb{v}^\text{cmp} = \text{AvgPool}(\text{ResBlock}(\mb{v}_t)) \nonumber 
\end{equation}

\paragraph{Sparse Reconstruction Transformer} 
The sparse reconstruction transformer uses the same number of transformer blocks (24) and hidden dimensions (1024) as the Stage 1 model. The key difference is that we replace the standard CrossAttn in Eq. \eqref{eq:crossattn_1} and \eqref{eq:crossattn_2} with NSACrossAttn, which uses GPU-friendly Triton kernels to combine coarse global context with sparse fine-grained context through separate attention branches. The compressed KV-head layout is crucial for making these Triton kernels memory efficient; Appendix~\ref{sec:app_nsa} provides additional implementation details. This structural alignment lets us initialize sparse-model weights directly from Stage 1, which significantly accelerates convergence. Furthermore, instead of predicting the high-resolution sparse volume from scratch, we train the sparse transformer to predict the residual over the dense prediction, an approach we found to be empirically much more effective. Let $\mb{x}_\text{d}^\text{up}$ and $\mb{y}_\text{d}^\text{up}$ denote the sparse tokens obtained by selecting and upsampling the informative regions of the dense predictions $\mb{x}_\text{d}$ and $\mb{y}_\text{d}$. The forward pass through the sparse transformer blocks is defined as:
\begin{align}
    \mb{x}^{(m)}, \mb{y}^{(m)} &= \text{NSABlock}^{(m)}\Big( \mb{x}^{(m-1)} + \text{Linear}^{(m)}(\mb{x}_\text{d}^\text{up}), \nonumber \\ 
    &\qquad\qquad\qquad\quad\; \mb{y}^{(m-1)} + \text{Linear}^{(m)}(\mb{y}_\text{d}^\text{up}) \Big), \quad \text{for } m = 1 \dots 24 \\ 
    \mb{x}_\text{s}, \mb{y}_\text{s} &= \mb{x}^{(24)} + \mb{x}_\text{d}^\text{up}, \, \mb{y}^{(24)} + \mb{y}_\text{d}^\text{up}
\end{align}
We project output tokens into a sparse feature volume $\mb{X}_\text{s}$ using the same linear layer as Eq. \eqref{eq:upsample}, resulting in resolution $S^\text{vol}_\text{f} = 4S^\text{vol}=384$. Note that $\mb{X}_\text{s}$ only contains features computed from selected active volume tokens. We therefore combine it with the full dense feature volume $\mb{X}_\text{d}$ during volume rendering. During ray marching, if a sampled point falls into a voxel $(i, j, k)$ where the mask $\mb{M}^\text{vol}_{i, j, k}$ is true, we query high-frequency features from $\mb{X}_\text{s}$. Conversely, if the point lies in empty space, we query the low-frequency features from $\mb{X}_\text{d}$. For points on the boundary between the two, we compute a linear combination of $\mb{X}_\text{s}$ and $\mb{X}_\text{d}$ to ensure smooth transitions. Finally, the lightweight MLP decoder processes these queried features exactly as in the dense reconstruction transformer.

\begin{figure}[t]
\begin{minipage}[c]{0.7\linewidth}
\includegraphics[width=\columnwidth]{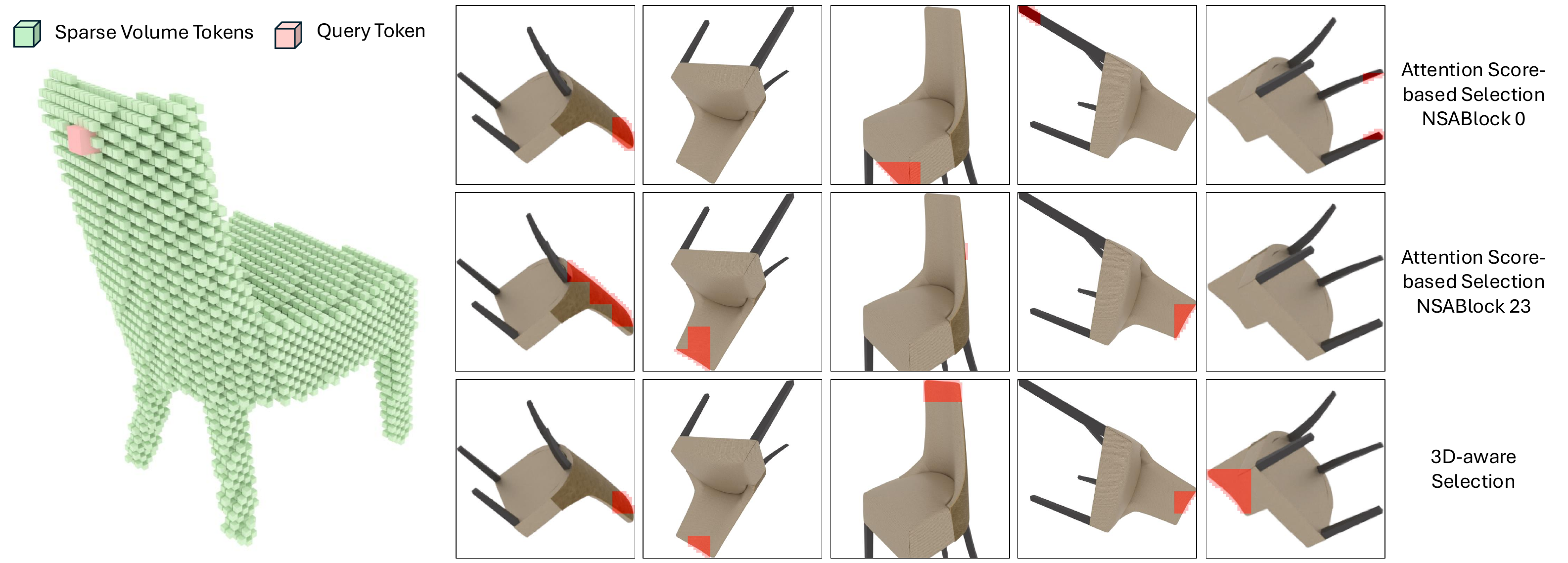}
\end{minipage}\hfill
\begin{minipage}[c]{0.28\linewidth}
\caption{\textbf{Attention-score vs.\ 3D-aware selection.} Standard attention captures 2D-3D correspondence only in deep layers, whereas our 3D-aware routing ensures stable selection to enhance texture.}
\label{fig:attn_vs_3daware}
\end{minipage}
\end{figure}

\subsection{3D-Aware Block Routing}
\label{sec:3daware}
Although our experiments show that LSRM benefits from an expanded context window and already improves fidelity over SOTA, standard NSA KV-block selection remains a bottleneck. In vanilla NSA, token blocks are chosen solely from attention scores against $\mb{k}^\text{cmp}$. As Fig. \ref{fig:attn_vs_3daware} illustrates, this data-driven routing often works in deeper layers, yet early transformer blocks frequently miss the most spatially relevant blocks, which cascades through the network and reduces final reconstruction quality.

To address this, we propose a 3D-aware routing strategy that selects KV-blocks using the coarse geometry established in Stage 1. We first assign an explicit 3D coordinate to each token. A volume token's coordinate $\mb{p}^\text{vol}$ is simply its voxel-center position. For an image token corresponding to an $8\times 8$ patch, its 3D location $\mb{p}^\text{img}$ is defined as the foreground surface point with the highest opacity, rendered via the dense feature volume $\mb{X}_\text{d}$. We then define the 3D point sets $\mathcal{P}^\text{img}$ and $\mathcal{P}^\text{vol}$ for image and volume blocks as the collections of token coordinates in the block. Our geometry-based block-selection rules are:

\paragraph{Image/Volume to Image Blocks} 
For a given query token with 3D coordinate $\mb{p} \in \{\mb{p}^\text{img}, \mb{p}^\text{vol}\}$, we first project $\mb{p}$ onto each input image plane using the known camera parameters. To limit computation, we retrieve the $B^\text{i}$ image blocks per view whose 2D centers are closest to this projection. Next, for each candidate block we compute the minimal 3D Euclidean distance between $\mb{p}$ and the point set $\mathcal{P}^\text{img}$. Finally, we select the $B^\text{i2i}$ (image queries) or $B^\text{v2i}$ (volume queries) candidates with the smallest such distances.

\paragraph{Image/Volume to Volume Blocks} 
Given the query coordinate $\mb{p} \in \{\mb{p}^\text{img}, \mb{p}^\text{vol}\}$, we compute its 3D Euclidean distance to the spatial centers of all volume token blocks. We then select the $B^\text{i2v}$ (for image queries) or $B^\text{v2v}$ (for volume queries) volume blocks with the smallest center distances.

Fig. \ref{fig:attn_vs_3daware} compares our \emph{volume-to-image} block selection against the original NSA mechanism on a converged model. The visualization confirms that our geometric approach establishes far more accurate 2D-3D correspondences. In Sec. \ref{sec:exp}, we show this enhanced spatially-aware routing translates to substantial improvements in reconstructed texture details (Fig. \ref{fig:gso_qual_abla} and Tab. \ref{tab:gso_quan}).

\begin{figure}[t]
\centering
\begin{minipage}[c]{0.7\linewidth}
\includegraphics[width=\columnwidth]{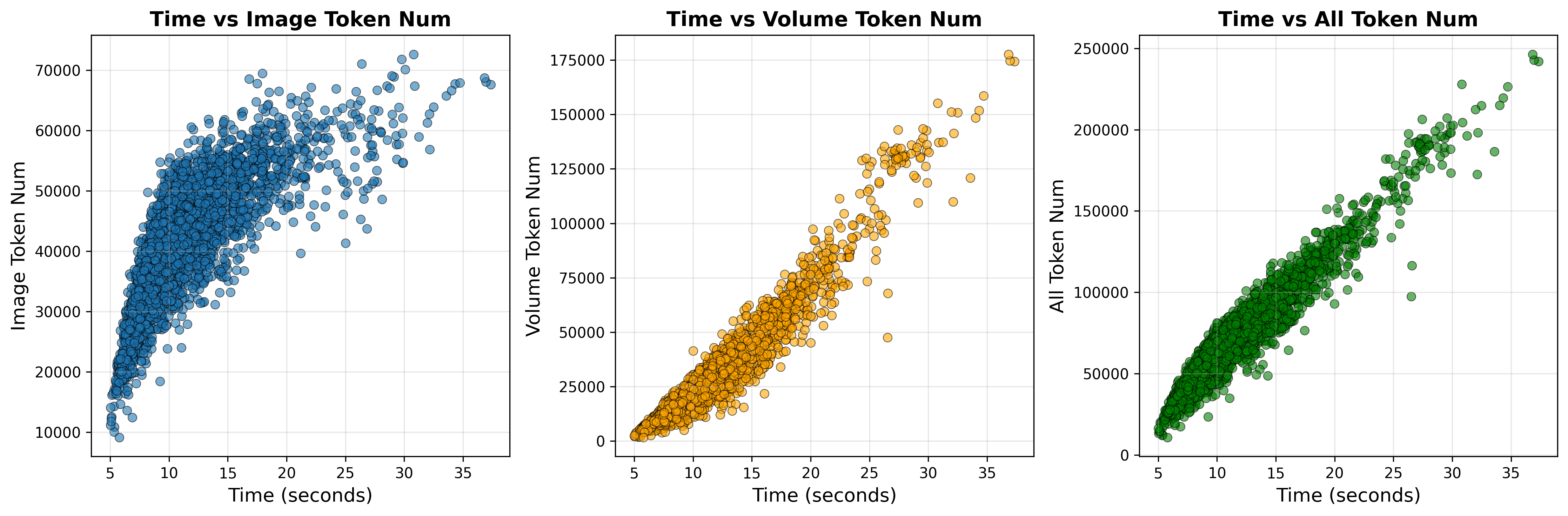}
\end{minipage}\hfill
\begin{minipage}[c]{0.28\linewidth}
\caption{\textbf{Token count vs. training time.} Distribution of active tokens and their corresponding training times for resolutions $S^\text{img}=96$ and $S^\text{vol}=64$. Each point represents a single data instance.}
\label{fig:time_token_dist}
\end{minipage}
\end{figure}

\subsection{Custom Sequence Parallelism for Workload Balancing} 
\label{sec:seqp}

\begin{figure}[t]
\centering
\includegraphics[width=\linewidth]{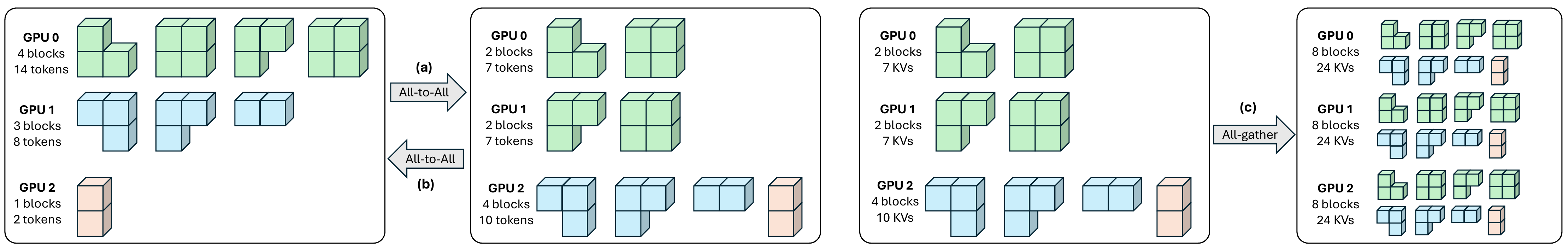}
\caption{\textbf{Custom sequence parallelism.} Toy example assuming $\le$4 tokens per spatial block across 3 GPUs. (a) Tokens are distributed evenly without breaking spatial blocks. (b) Tokens return to their source GPUs for decoding and volume rendering. (c) All-gather-KV ensures global context before every NSACrossAttn layer.}
\label{fig:seqp}
\end{figure}

LSRM’s sparse tokenization yields highly variable active-token counts across instances, producing straggler GPUs that stall training (Fig. \ref{fig:time_token_dist}) and tightly limit attainable resolution during multi-GPU training, since each step waits for the slowest worker. Early attempts to cap per-GPU tokens with reservoir sampling routinely dropped as many as one third of active tokens, which destabilized optimization. To balance this dynamic workload systematically, we adapt context parallelism \cite{korthikanti2023reducing} into a custom \textbf{block-aware sequence parallelism}. Unlike naive splits that break the block structure required by NSA, our scheme spreads tokens evenly across GPUs while strictly keeping each spatial block on a single device. This preserves block locality, so window attention and KV-block compression are computed entirely on-device, with zero cross-GPU communication.

To execute compressed (CmpAttn) and selected (SelAttn) attention over global context, tokens must still be routed efficiently. We use an All-gather-KV strategy \cite{liu2023ring,fang2024usp} that exploits our high KV compression. Concretely, we all-gather to replicate $(\mb{k}^\text{cmp}, \mb{v}^\text{cmp})$ and $(\mb{k}, \mb{v})$ on every GPU, while keeping queries ($\mb{q}$) locally sharded. Since the KV cache is small, the communication cost of this gather is modest. The result is a favorable trade-off that lets us balance compute across 8 GPUs. As shown in Fig. \ref{fig:seqp}, we perform an all-to-all collective before the first NSABlock to distribute token blocks globally, and then invoke all-gather on keys and values before SelAttn and CmpAttn.

\subsection{Implementation, Training, and Inference Details}
We train LSRM for novel-view synthesis (NVS) on 600K curated 3D models and then fine-tune it for inverse rendering. At each iteration, we randomly select 12--16 input views and 4 target views from 32 camera poses sampled on a sphere, following LIRM \cite{li2025lirm} and MeshLRM \cite{wei2024meshlrm} to ensure robustness to diverse camera trajectories. NVS training follows three stages: (1) We train the dense reconstruction transformer ($S^{\text{vol}}_{\text{d}}=16, S^{\text{img}}_{\text{d}}=32$), yielding a decoded volume resolution of $S^{\text{vol}}_{\text{df}}=64$ for $256 \times 256$ inputs. (2) We freeze the dense model and use its predictions to train the sparse reconstruction transformer on the fly. We start at lower resolutions ($S^{\text{img}}=64, S^{\text{vol}}=64$) to support larger batch sizes and quicker convergence. (3) We scale sparse resolutions to $S^{\text{img}}=96, S^{\text{vol}}=96$. This final stage relies on our custom block-aware sequence parallelism for stable training, with active image and volume token counts peaking at roughly 100K and 300K, respectively. Compared to recent SOTA like LIRM \cite{li2025lirm}---which processes a maximum of 8 images ($64 \times 64 \times 8 \approx 32\text{K}$ image tokens) and uses a hexaplane representation ($48 \times 48 \times 6 \approx 14\text{K}$ object tokens)---LSRM scales to $>\mathbf{2\times}$ more image tokens and $>\mathbf{20\times}$ more object tokens. These three stages require 5, 7, and 3 days, respectively, on 128 H100 GPUs. Across all stages, we optimize rendering, foreground mask, and depth losses, adding a numerical normal loss \cite{li2025lirm} during the final 500--1000 iterations to boost geometric fidelity.

Because our goal is to evaluate context-window scaling, we reduce compute by initializing inverse rendering from the converged NVS weights. We apply two task-specific architectural changes: (1) we encode background images with a linear layer \cite{li2025lirm} to better disentangle lighting from materials, and (2) we initialize the BRDF prediction head with pre-trained weights from $\textbf{MLP}_\text{c}$. We reuse the same 600K 3D models, rendering them under real HDR environment maps with extensive augmentations. Fine-tuning mirrors the three-stage NVS schedule. For efficiency, we initialize the first two stages directly from their NVS counterparts and halve the training iterations, while the final stage initializes from the converged second stage. Additional training and loss details are provided in Appendix~\ref{sec:app_details}. 

All inference experiments are conducted on a single NVIDIA H100 GPU. With 16 input images at the highest resolution, LSRM requires 40 GB peak GPU memory, and the sparse reconstruction transformer reconstructs an object in about 5 seconds on average. This remains orders of magnitude faster than optimization-based alternatives \cite{sun2023neural,munkberg2022extracting} and is comparable to LIRM's iterative refinement when using the same number of views.

%% file: section/experiments.tex
\section{Experiments}
\label{sec:exp}

\begin{figure}[t]
    \centering
    \includegraphics[width=\linewidth]{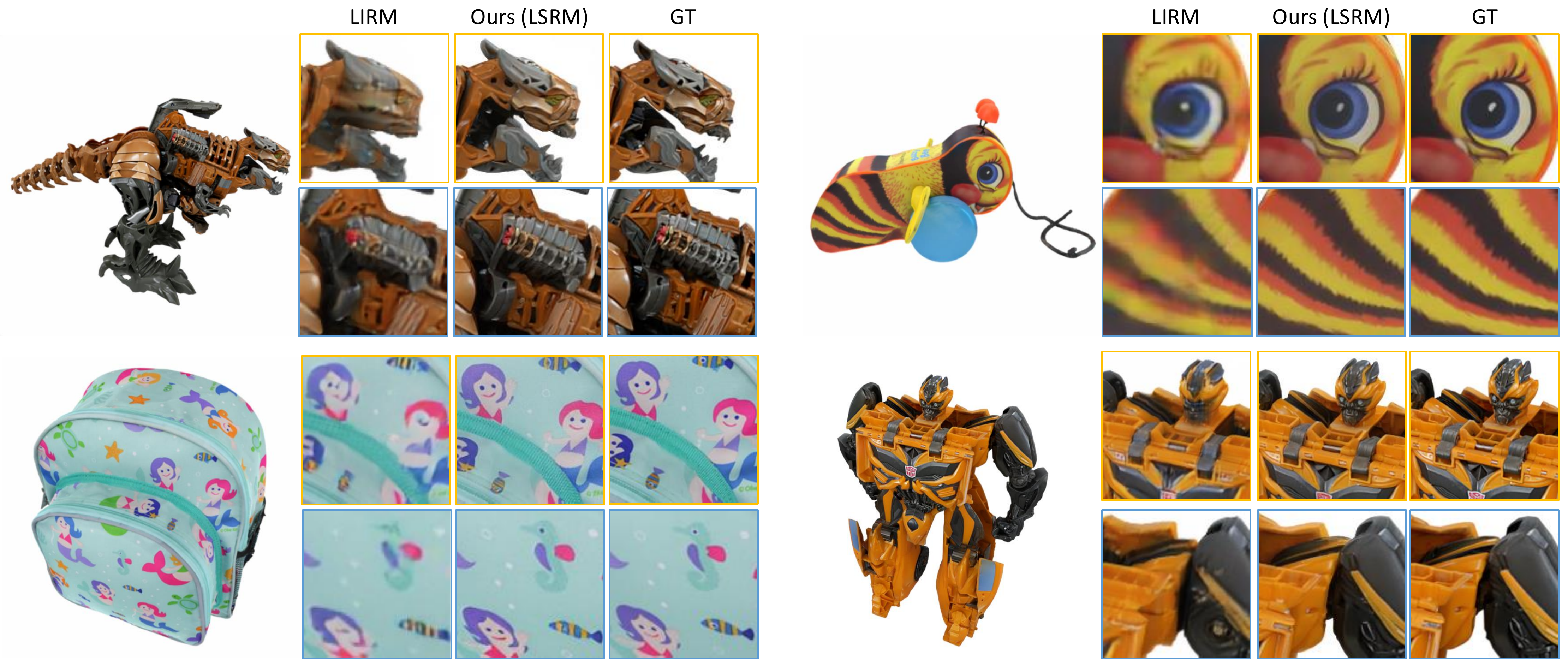}
\caption{\textbf{Qualitative NVS Results on the GSO Dataset.} LSRM significantly improves texture fidelity over prior SOTA \cite{li2025lirm}. Zoomed-in comparisons show that our method successfully recovers legible text (bottom left), complex geometry (top left) and sharp facial structures (right), whereas the baseline produces blurred artifacts.}
\label{fig:gso_qual}
\end{figure}

\paragraph{Object-centric Novel View Synthesis}

\begin{figure}[t]
    \centering
    \includegraphics[width=\linewidth]{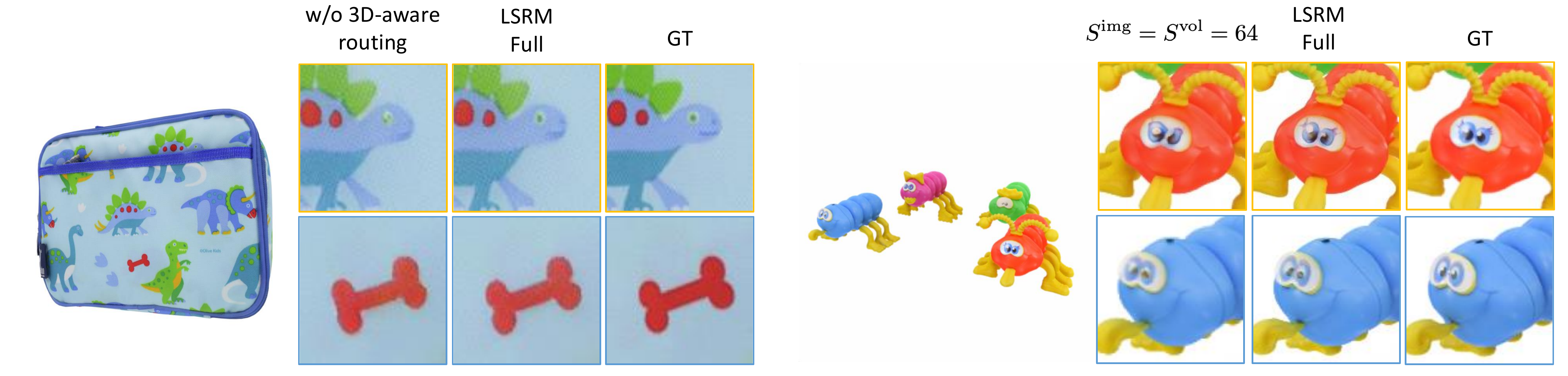}
    \caption{\textbf{Qualitative Ablation Study on the GSO Dataset.} Visual comparisons demonstrate that both our 3D-aware block routing and increased spatial resolutions independently enhance texture fidelity and overall rendering quality.}
    \label{fig:gso_qual_abla}
\end{figure}

We evaluate the NVS quality of LSRM on the GSO dataset \cite{downs2022google}. Inputs are 16 uniformly sampled 768 × 768 views rendered with the camera configuration of \cite{li2025lirm}, and we render 12 768 × 768 novel views for evaluation. Following standard practice, ground-truth novel views are rendered at 512×512, and we downsample our predictions to this resolution before computing metrics. We compute PSNR, SSIM and LPIPS on the rendered RGB images, averaging over all objects and target views, and we use identical evaluation scripts across methods. Quantitative and qualitative results are summarized in Tab. \ref{tab:gso_quan} and Fig. \ref{fig:gso_qual}, respectively. In Tab. \ref{tab:gso_quan}, we compare LSRM against state-of-the-art feed-forward baselines. LIRM \cite{li2025lirm} uses the same number of input images via iterative refinement, but operates at a lower 512 × 512 resolution. By contrast, LSRM increases the active image and volume token counts during reconstruction, yielding a substantial boost in overall fidelity. Our full model improves PSNR by > 2.4 dB and reduces LPIPS by > 40\% relative to the strongest baseline. These gains are also evident qualitatively: LSRM produces sharper textures and recovers recognizable facial structure (e.g., the Little Mermaid and the bee). Even for objects with complex geometry (transformers, top left and bottom right), zoomed-in crops are nearly indistinguishable from ground truth. We additionally perform ablations to validate key components, including context-window scaling via $S^\text{img}$ and $S^\text{vol}$, as well as our 3D-aware block routing. Results in Tab. \ref{tab:gso_quan} and Fig. \ref{fig:gso_qual_abla} show that enlarging the context window provides the largest gains, while 3D-aware routing further sharpens fine-grained structures and restores previously distorted faces. Separately scaling the image-token and volume-token context lengths with 3D-aware routing shows that both contribute to reconstruction quality, with volume-token scaling having the larger impact: reducing $S^\text{vol}$ from 64 to 48 drops PSNR to 31.68, compared to 31.82 when reducing $S^\text{img}$ from 64 to 48.

\begin{table}[t]
\centering
\caption{\textbf{Quantitative NVS Results on the GSO Dataset.} We compare LSRM against SOTA feed-forward baselines \cite{wei2024meshlrm,zhang2025gs, li2025lirm}. Our ablations evaluate the dense representation, context-window scaling through the image and volume token resolutions $S^\text{img}$ and $S^\text{vol}$, and 3D-aware block routing. Best, second best, and third best results are highlighted in \textcolor{red}{\textbf{red}}, \textcolor{orange}{orange}, and \textcolor{yellow!90!orange}{yellow}, respectively.}
\label{tab:gso_quan}
\setlength{\tabcolsep}{6pt}  
{
\setlength{\tabcolsep}{1pt}
\fontsize{6pt}{7pt}\selectfont
\begin{tabular}{r ccc ccccccc}
\toprule
& \multicolumn{3}{c}{\textbf{Baselines}} & \multicolumn{7}{c}{\textbf{Ours (LSRM Ablations)}} \\
\cmidrule(lr){2-4} \cmidrule(lr){5-11}
\multirow{2}{*}{\textbf{Metric}} & Mesh & GS & \multirow{2}{*}{LIRM} & \multirow{2}{*}{Dense} & $S^\text{img}\!=\!64$ & $S^\text{img}\!=\!48$ & \multicolumn{2}{c}{$S^{\text{img}}\!=\!S^{\text{vol}}=64$} & \multicolumn{2}{c}{$S^{\text{img}}\!=\!S^{\text{vol}}=96$} \\
\cmidrule(lr){8-9} \cmidrule(lr){10-11}
& -LRM & -LRM &  & model & $S^\text{vol}\!=\!48$ & $S^\text{vol}\!=\!64$ & w/o routing & w/ routing & w/o routing & w/ routing \\
\midrule
\textbf{PSNR ($\uparrow$)}    & 28.13 & 30.52 & 30.65 & 28.45 & 31.68 & 31.82 & 31.34 & \cellcolor{yellow!15}31.93 & \cellcolor{orange!15}32.72 & \cellcolor{red!15}\textbf{33.08} \\
\textbf{SSIM ($\uparrow$)}    & 0.923 & 0.952 & 0.949 & 0.934 & 0.958 & 0.959 & 0.956 & \cellcolor{yellow!15}0.962 & \cellcolor{orange!15}{0.968} & \cellcolor{red!15}\textbf{0.971} \\
\textbf{LPIPS ($\downarrow$)} & 0.093 & 0.050 & 0.054 & 0.079 & 0.044 & 0.043 & 0.045 & \cellcolor{yellow!15}0.040 & \cellcolor{orange!15}0.032 & \cellcolor{red!15}\textbf{0.028} \\
\bottomrule
\end{tabular}
}
\end{table}

\begin{table}[t]
\centering
\caption{\textbf{Quantitative Inverse Rendering Results.} We evaluate our approach against optimization-based and feed-forward baselines across three benchmarks. Notably, LSRM consistently achieves the lowest LPIPS across all datasets, a metric highly sensitive to fine texture quality and perceptual fidelity.}
\label{tab:inv_render_quan}
{
\setlength{\tabcolsep}{1pt}
\fontsize{6pt}{7pt}\selectfont
\begin{tabular}{l ccccc cccc ccc}
\toprule
\multirow{2}{*}{\textbf{Method}} & \multicolumn{5}{c}{\textbf{StanfordORB} \cite{kuang2024stanford}} & \multicolumn{4}{c}{\textbf{DTC} \cite{dong2025digital}} & \multicolumn{3}{c}{\textbf{OWL} \cite{ummenhofer2024objects}} \\
\cmidrule(lr){2-6} \cmidrule(lr){7-10} \cmidrule(lr){11-13}
& \shortstack{PSNR-H\\$\uparrow$}
& \shortstack{PSNR-L\\$\uparrow$}
& \shortstack{SSIM\\$\uparrow$}
& \shortstack{LPIPS\\$\downarrow$}
& \shortstack{CD\\$\downarrow$}
& \shortstack{PSNR-H\\$\uparrow$}
& \shortstack{PSNR-L\\$\uparrow$}
& \shortstack{SSIM\\$\uparrow$}
& \shortstack{LPIPS\\$\downarrow$}
& \shortstack{PSNR\\$\uparrow$}
& \shortstack{SSIM\\$\uparrow$}
& \shortstack{LPIPS\\$\downarrow$} \\
\midrule
NVDiffrecMc             & 24.43 & 31.60 & \cellcolor{yellow!15}0.972 & 0.036 & 0.51 & \cellcolor{yellow!15}27.78 & 34.55 & 0.952 & 0.042 & 19.82 & 0.73 & 0.389 \\
InvRender               & 23.76 & 30.83 & 0.970 & 0.046 & 0.44 & \cellcolor{orange!15}29.52 & \cellcolor{red!15}\textbf{35.98} & \cellcolor{orange!15}0.961 & \cellcolor{yellow!15}0.037 & \cellcolor{orange!15}23.77 & \cellcolor{orange!15}0.78 & \cellcolor{yellow!15}0.369 \\
NeuralPBIR              & \cellcolor{red!15}\textbf{26.01} & \cellcolor{red!15}\textbf{33.26} & \cellcolor{red!15}\textbf{0.979} & \cellcolor{orange!15}0.023 & \cellcolor{yellow!15}0.43 &  N/A  &  N/A  &  N/A  &  N/A  &  N/A  &  N/A &  N/A  \\
LIRM                    & \cellcolor{yellow!15}25.09 & \cellcolor{yellow!15}32.45 & \cellcolor{yellow!15}0.972 & \cellcolor{yellow!15}0.025 & \cellcolor{orange!15}0.38 & 27.65 & \cellcolor{yellow!15}34.84 & \cellcolor{yellow!15}0.960 & \cellcolor{orange!15}0.031 & \cellcolor{yellow!15}23.27 & \cellcolor{yellow!15}0.77 & \cellcolor{orange!15}0.322 \\
\textbf{Ours (LSRM)}    & \cellcolor{orange!15}25.47 & \cellcolor{orange!15}32.85 & \cellcolor{orange!15}0.977 & \cellcolor{red!15}\textbf{0.022} & \cellcolor{red!15}\textbf{0.29} & \cellcolor{red!15}\textbf{29.67} & \cellcolor{orange!15}35.66 & \cellcolor{red!15}\textbf{0.964} & \cellcolor{red!15}\textbf{0.025} & \cellcolor{red!15}\textbf{24.88} & \cellcolor{red!15}\textbf{0.79} & \cellcolor{red!15}\textbf{0.269} \\
\bottomrule
\end{tabular}
}
\end{table}

\begin{figure}[t]
\centering
\includegraphics[width=\linewidth]{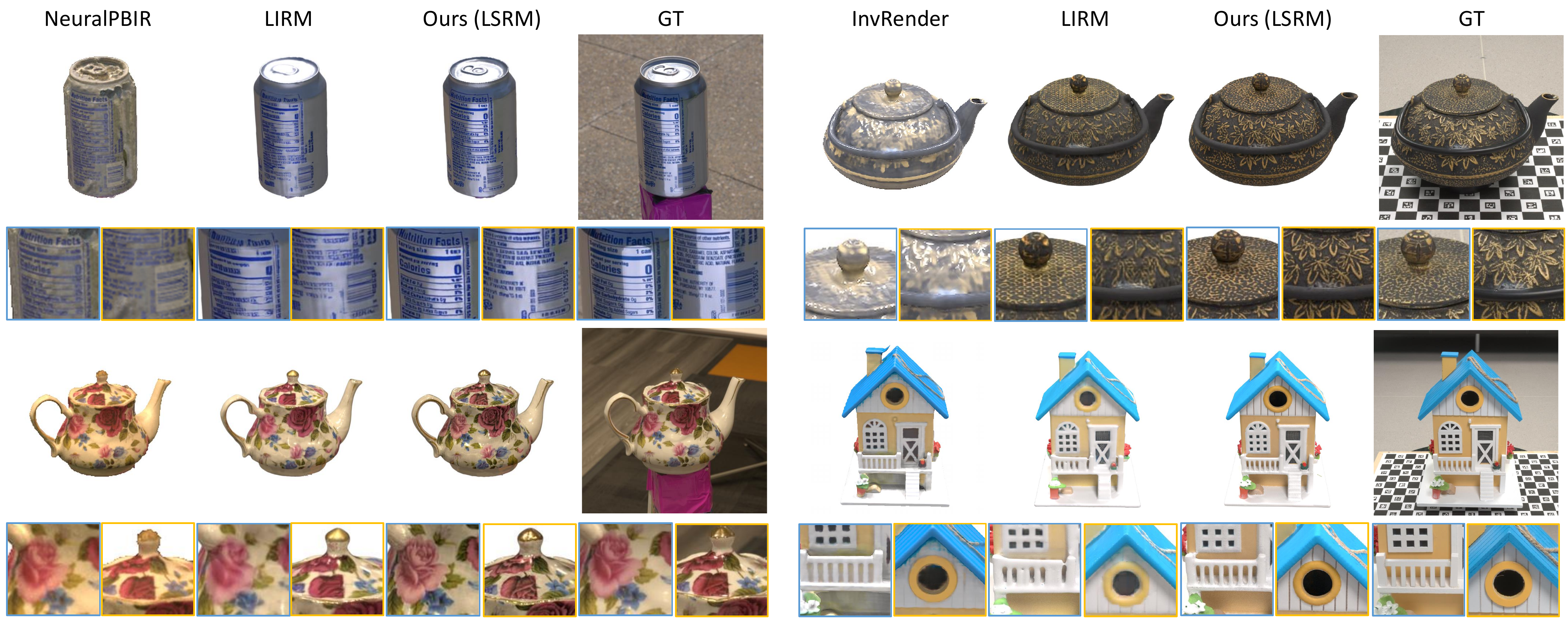}
\caption{\textbf{Qualitative Inverse Rendering Results.} Visual comparisons against optimization-based and feed-forward baselines demonstrate that LSRM consistently recovers higher-fidelity textures and finer geometric details.}
\label{fig:inv}
\end{figure}

\begin{figure}[t]
    \centering
    \includegraphics[width=\linewidth]{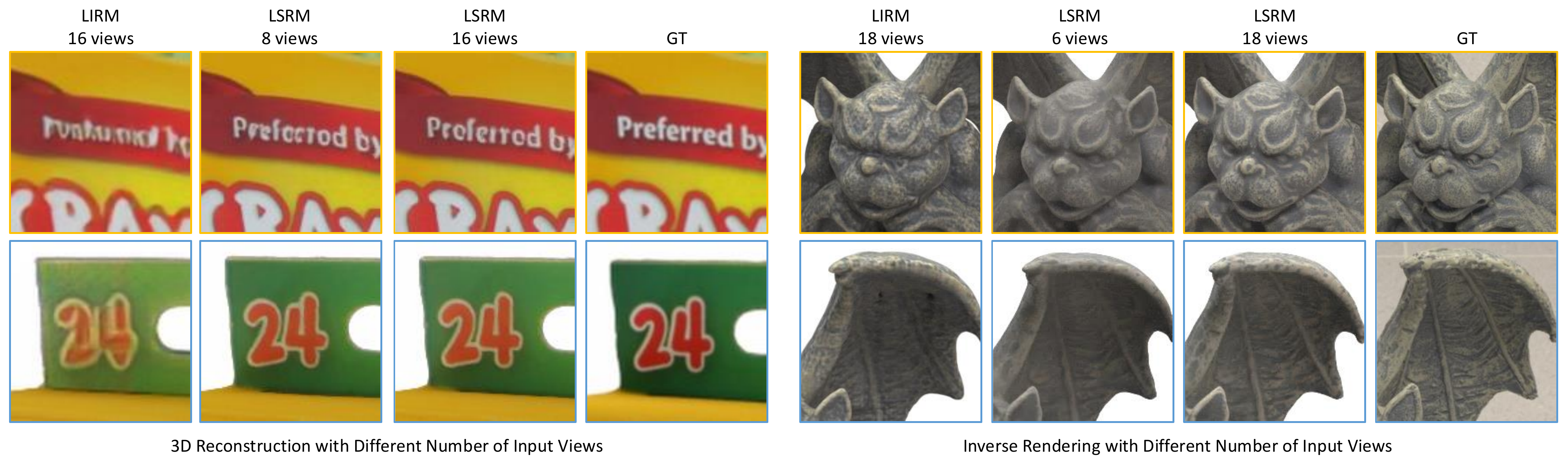}
    \caption{\textbf{Qualitative Results with Fewer Input Views.} LSRM preserves high-fidelity geometry and texture when evaluated with reduced-view inputs for both novel-view synthesis and inverse rendering, even though it is trained with 12--16 views.}
    \label{fig:qual_fewviews}
\end{figure}

\begin{table}[t]
    \centering
    \begin{minipage}[c]{0.43\linewidth}
        {
        \setlength{\tabcolsep}{1pt}  
        \fontsize{6pt}{7pt}\selectfont
        \begin{tabular}{r cc cc}
            \toprule
            \multirow{2}{*}{\textbf{Metric}} & \multicolumn{2}{c}{\textbf{LIRM} \cite{li2025lirm}} & \multicolumn{2}{c}{\textbf{Ours (LSRM)}} \\
            \cmidrule(lr){2-3} \cmidrule(lr){4-5}
            & 8 views & 16 views & 8 views & 16 views \\
            \midrule
            \textbf{PSNR ($\uparrow$)}    & 30.48 & \cellcolor{yellow!15}30.56 & \cellcolor{orange!15}32.46 & \cellcolor{red!15}\textbf{33.08} \\
            \textbf{SSIM ($\uparrow$)}    & 0.947 & \cellcolor{yellow!15}0.948 & \cellcolor{orange!15}0.968 & \cellcolor{red!15}\textbf{0.971} \\
            \textbf{LPIPS ($\downarrow$)} & 0.056 & \cellcolor{yellow!15}0.054 & \cellcolor{orange!15}0.031 & \cellcolor{red!15}\textbf{0.028} \\
            \bottomrule
        \end{tabular}
        }
    \end{minipage}\hfill
    \begin{minipage}[c]{0.55\linewidth}
        \caption{\textbf{Quantitative NVS Results with Fewer Input Views.} On GSO \cite{downs2022google}, LSRM maintains strong performance with 8 input views and substantially outperforms LIRM using either 8 or 16 views.}
        \label{tab:gso_quan_fewviews}
    \end{minipage}
\end{table}

\begin{table}[th!]
\centering
\caption{\textbf{Quantitative Inverse Rendering Results with Fewer Input Views.} With only 6 input views, LSRM outperforms feed-forward relighting and inverse-rendering baselines across the evaluated benchmarks.}
\label{tab:inv_render_quan_fewviews}
{
\setlength{\tabcolsep}{1pt}
\fontsize{6pt}{7pt}\selectfont
\begin{tabular}{c l ccccc cccc ccc}
\toprule
\multirow{2}{*}{\textbf{Views}} & \multirow{2}{*}{\textbf{Method}} & \multicolumn{5}{c}{\textbf{StanfordORB} \cite{kuang2024stanford}} & \multicolumn{4}{c}{\textbf{DTC} \cite{dong2025digital}} & \multicolumn{3}{c}{\textbf{OWL} \cite{ummenhofer2024objects}} \\
\cmidrule(lr){3-7} \cmidrule(lr){8-11} \cmidrule(lr){12-14}
&
& \shortstack{PSNR-H\\$\uparrow$}
& \shortstack{PSNR-L\\$\uparrow$}
& \shortstack{SSIM\\$\uparrow$}
& \shortstack{LPIPS\\$\downarrow$}
& \shortstack{CD\\$\downarrow$}
& \shortstack{PSNR-H\\$\uparrow$}
& \shortstack{PSNR-L\\$\uparrow$}
& \shortstack{SSIM\\$\uparrow$}
& \shortstack{LPIPS\\$\downarrow$}
& \shortstack{PSNR\\$\uparrow$}
& \shortstack{SSIM\\$\uparrow$}
& \shortstack{LPIPS\\$\downarrow$} \\
\midrule
\multirow{3}{*}{\textbf{6}} 
& RelitLRM             & \cellcolor{yellow!15}24.67 & \cellcolor{yellow!15}31.52 & \cellcolor{yellow!15}0.969 & \cellcolor{yellow!15}0.032 & N/A & N/A & N/A & N/A & N/A & \cellcolor{orange!15}23.08 & \cellcolor{orange!15}0.79 & \cellcolor{orange!15}0.284 \\
& LIRM                 & \cellcolor{orange!15}24.76 & \cellcolor{orange!15}32.11 & \cellcolor{orange!15}0.971 & \cellcolor{orange!15}0.027 & \cellcolor{orange!15}0.48 & \cellcolor{orange!15}28.03 & \cellcolor{orange!15}34.28 & \cellcolor{orange!15}0.959 & \cellcolor{orange!15}0.034 & \cellcolor{yellow!15}22.70 & \cellcolor{yellow!15}0.76 & \cellcolor{yellow!15}0.326 \\
& \textbf{Ours} & \cellcolor{red!15}\textbf{25.57} & \cellcolor{red!15}\textbf{32.74} & \cellcolor{red!15}\textbf{0.975} & \cellcolor{red!15}\textbf{0.023} & \cellcolor{red!15}\textbf{0.32} & \cellcolor{red!15}\textbf{29.09} & \cellcolor{red!15}\textbf{35.16} & \cellcolor{red!15}\textbf{0.961} & \cellcolor{red!15}\textbf{0.028} & \cellcolor{red!15}\textbf{24.43} & \cellcolor{red!15}\textbf{0.80} & \cellcolor{red!15}\textbf{0.255} \\
\midrule
\textbf{18} & \textbf{Ours} & 25.47 & 32.85 & 0.978 & 0.022 & 0.29 & 29.67 & 35.66 & 0.965 & 0.025 & 24.88 & 0.79 & 0.269 \\
\bottomrule
\end{tabular}
}
\end{table}

\paragraph{Object-centric Inverse Rendering}  We evaluate LSRM’s inverse-rendering extension on three widely used benchmarks: StanfordORB \cite{kuang2024stanford}, DigitalTwinCatalogue (DTC) \cite{dong2025digital}, and ObjectsWithLighting (OWL) \cite{ummenhofer2024objects}. These datasets span diverse materials, illumination, and capture setups, offering a stringent test of relightability, texture fidelity, and geometric consistency. We focus our comparisons on optimization-based and feed-forward baselines with strong reported quantitative results or available qualitative comparisons. As summarized in Tab.~\ref{tab:inv_render_quan}, LSRM consistently outperforms prior feed-forward baselines. However, gains in pixel-wise metrics such as PSNR are comparatively modest. This reflects the inherently ill-posed nature of inverse rendering: although LSRM recovers high-fidelity geometry and texture details, global ambiguities—e.g., slight global shifts in albedo, roughness and metallicity maps—can disproportionately penalize PSNR. In contrast, on perceptual metrics like LPIPS, which emphasize high-frequency structure, LSRM shows substantial improvements, reaching performance comparable to SOTA optimization-based methods.

Qualitative results in Fig.~\ref{fig:inv} corroborate these gains in texture fidelity. For example, LSRM recovers legible text (the ``Nutrition Facts'' and ``Calories 0'' label on the can, top right) and fine textures (the golden leaves and flowers on the teapot, top right and bottom left). Beyond appearance, LSRM also improves geometry, accurately recovering the small fence in front of the birdhouse. This geometric accuracy is reflected in StanfordORB Chamfer Distance (CD), which drops markedly relative to LIRM and optimization-based approaches. 

\paragraph{Generalization to Smaller Number of Views}
Since the primary focus of LSRM is to significantly scale the context window for both image and volume tokens, we maintain the number of input images between 12 and 16 during training and do not fine-tune our models on fewer images. Surprisingly, both our feed-forward 3D reconstruction and inverse rendering models generalize well to a smaller number of input views, outperforming the prior state-of-the-art, LIRM \cite{li2025lirm}, which was explicitly trained on 4 to 8 images. We evaluate this generalization capability for novel-view synthesis on the GSO dataset \cite{downs2022google}. As summarized in Tab. \ref{tab:gso_quan_fewviews} and Fig. \ref{fig:qual_fewviews}, providing fewer views causes only a slight drop in texture sharpness compared to the full 16-view input. However, our results remain substantially superior to LIRM whether using 8 or 16 input views, as reflected in the quantitative metrics. Specifically, our 8-view results achieve close to a \textbf{2 dB} increase in PSNR compared to the 16-view LIRM, which corresponds to an approximate \textbf{37\%} drop in MSE loss, while the LPIPS loss drops by more than \textbf{40\%}. We observe a similar phenomenon in inverse rendering, where LSRM consistently outperforms LIRM using only 6 input views. Furthermore, evaluating on fewer views enables a fair comparison with RelitLRM \cite{zhang2024relitlrm}, another feed-forward relighting model. Unlike our approach, RelitLRM does not explicitly predict material reflectance; rather, it relies on a generative network to synthesize novel lighting appearances, resulting in slower inference. Nevertheless, LSRM outperforms both RelitLRM and LIRM across two popular relighting benchmarks (see Tab. \ref{tab:inv_render_quan_fewviews}), with qualitative results in Fig. \ref{fig:qual_fewviews} corroborating these improvements.

%% file: section/conclusions.tex
\section{Conclusions}
\label{sec:conclusions}

In this work, we show that scaling transformer context windows markedly improves feed-forward 3D reconstruction and inverse rendering. By scaling object and image tokens by $20\times$ and $2\times$ with NSA, we close much of the texture- and geometry-fidelity gap to optimization. To make this scale practical, we introduce 3D-aware block routing and block-aware sequence parallelism, boosting rendering quality while improving GPU utilization. Extensive experiments demonstrate that LSRM sets a new SOTA, recovering crisp text and faithful facial structures.

\paragraph{Limitations and Future Work}
LSRM introduces a general coarse-to-fine pipeline that leverages NSA to effectively scale context windows for 3D-related tasks. While this paper validates its effectiveness on object-centric 3D reconstruction and inverse rendering, we hope to extend this framework to large-scale scene reconstruction and general video/image generation in future work. Despite achieving SOTA reconstruction quality, several limitations remain. First, we observe that LSRM does not inherently resolve other ill-posed challenges in inverse rendering, such as accurately estimating roughness and metallic material properties. Furthermore, we still encounter cases where extremely fine details—such as small ingredient lists on cans—remain illegible. To address this, future iterations could expand the context window even further by integrating sequence parallelism techniques like Ulysses attention \cite{jacobs2023deepspeed} and ring attention \cite{liu2023ring}.

\paragraph{Acknowledgements} We would like to thank David Clabaugh for creating the teaser figures, Dilin Wang and Yuchen Fan for their support with the training infrastructure, and Hao Tan, Sai Bi, and Kai Zhang for technical discussions regarding sparse attention.

%% file: section/nsa.tex
\section{Background: Native Sparse Attention}
\label{sec:app_nsa}

NSA is a hardware-friendly mechanism designed for highly efficient end-to-end training and inference on modern NVIDIA GPUs. We adopt NSA to scale the context window because, unlike earlier sparse-attention schemes that impose fixed block-to-block patterns \cite{zaheer2020big,child2019generating}, NSA lets each query adaptively choose the sparse KV-blocks it attends to, which we find essential for high-fidelity reconstruction. Let $\mb{q} \in \mathbb{R}^{N_x \times h_q \times d_h}$ and $\mb{k}, \mb{v} \in \mathbb{R}^{N_y \times h_{kv} \times d_h}$ denote the query, key, and value tensors computed from the input token sequences $\mb{x} \in \mathbb{R}^{N_x \times d}$ and $\mb{y} \in \mathbb{R}^{N_y \times d}$, respectively. Note that in self-attention, $\mb{x}$ and $\mb{y}$ are the same tensor. Here, $N_x$ and $N_y$ represent the sequence lengths, $h_q$ and $h_{kv}$ denote the number of heads for queries and keys/values, and $d_h$ is the head dimension, so $d = h_q \times d_h$. NSA dynamically selects and recomputes keys $\mb{k}_i^{(\cdot)}$ and values $\mb{v}_i^{(\cdot)}$ for each query $\mb{q}_i \in \mathbb{R}^{h_q \times d_h}$ using specific strategies based on original $\mb{q}$, $\mb{k}$, and $\mb{v}$. The slightly simplified variant employed in LSRM is formulated as follows:

\paragraph{Compressed} ($\mb{k}^\text{cmp}, \mb{v}^\text{cmp}$): We divide $\mb{k}$ and $\mb{v}$ into $B$ non-overlapping blocks, where $B \ll N_y$. The keys and values within each block are then compressed into a single key-value pair $\mb{k}^\text{cmp}, \mb{v}^\text{cmp}$. Note that because $\mb{k}^\text{cmp}$ and $\mb{v}^\text{cmp}$ are identical for all queries, we omit the subscript $i$ for them.

\paragraph{Selected} ($\mb{k}_i^\text{sel}, \mb{v}_i^\text{sel}$): For each query $\mb{q}_i$, we compute its attention scores against the compressed keys $\mb{k}^\text{cmp}$ and select the top $B^\text{sel}$ KV-blocks (with $B^\text{sel} \ll B$). We then define $\mb{k}^\text{sel}_i, \mb{v}^\text{sel}_i$ as the set of all original keys and values contained within these $B^\text{sel}$ selected blocks.

\paragraph{Window} ($\mb{k}_i^\text{win}, \mb{v}_i^\text{win}$): For each query $\mb{q}_i$, we define $\mb{k}^\text{win}_i, \mb{v}^\text{win}_i$ as all keys and values located in the same block as the original $\mb{k}_i$ and $\mb{v}_i$.  

Our NSACrossAttn layer outputs a gated combination of these three attention branches. Let $\mb{o}_i$ denote the output for query $\mb{q}_i$:
\begin{align}
\mb{o}_i &= \mb{w}_i^{\text{cmp}} \text{CmpAttn}(\mb{q}_i, \mb{k}^\text{cmp}, \mb{v}^\text{cmp}) + \mb{w}_i^{\text{sel}} \text{SelAttn}(\mb{q}_i, \mb{k}^\text{sel}_i, \mb{v}^\text{sel}_i) \nonumber \\
&\quad + \mb{w}_i^{\text{win}} \text{WinAttn}(\mb{q}_i, \mb{k}^\text{win}_i, \mb{v}^\text{win}_i) \label{eq:nsa_1} \\ 
\mb{w}_i^{\text{cmp}}, \mb{w}_i^{\text{sel}}, \mb{w}_i^{\text{win}} &= \text{Sigmoid}(\text{Linear}_{\text{gate}}(\mb{x}_i)) \label{eq:nsa_2}
\end{align}
Following Qiu et al. \cite{qiu2025gated}, we predict gating weights $\mb{w}_i^{\text{cmp}}, \mb{w}_i^{\text{sel}}, \mb{w}_i^{\text{win}}$ with the same dimension as $\mb{x}_i$ to ensure more stable training. These three attention branches are highly complementary: $\text{CmpAttn}$ efficiently captures global context due to its reduced sequence length ($B \ll N_y$), $\text{SelAttn}$ retrieves fine-grained details from the most relevant regions with low computational cost ($B^\text{sel} \ll B$), and $\text{WinAttn}$ models strictly local context with minimal overhead. 

Among the three branches, $\text{CmpAttn}$ and $\text{WinAttn}$ can be directly implemented with FlashAttn \cite{dao2023flashattention}. $\text{SelAttn}$, however, requires a custom Triton kernel \cite{tillet2021triton} because each query $\mb{q}_i$ attends to a dynamically chosen sparse set of KV-blocks. Instead of loading a full block of queries and KV pairs into SRAM, NSA leverages the high compression ratio of GQA \cite{ainslie2023gqa} to load multiple query heads together with a single shared head of $\mb{k}^{\text{sel}}_i$ and $\mb{v}^{\text{sel}}_i$. To fully utilize NVIDIA Tensor Cores, the ratio $h_q / h_{kv}$ should be a multiple of \textbf{16} \cite{lai2025nativesparseattention}. In LSRM, we build on the PyTorch reference implementation \cite{wang2025nativesparseattentionpytorch} and adapt the Triton implementation \cite{lai2025nativesparseattention}, setting $h_q = 32$ and $h_{kv} = 2$. This high ratio also enables more efficient sequence parallelism, as discussed in Sec.~\ref{sec:seqp}.

%% file: section/details.tex
\section{Additional Training Details and Loss Functions}
\label{sec:app_details}
We optimize our model with AdamW \cite{loshchilov2018decoupled}, using $\beta_1=0.9$ and $\beta_2=0.95$. To improve training stability, we apply weight normalization and QK normalization \cite{henry2020query}. During training, we use Fully Sharded Data Parallel (FSDP) in fully sharded mode, wrapping the dense reconstruction transformer, the sparse reconstruction transformer, and the frozen DINOv3 \cite{simeoni2025dinov3} encoder to reduce GPU memory consumption. For NSA, we keep the block sizes fixed at $8^3$ for volume tokens and $8^2$ for image tokens across all resolutions, while adjusting the number of selected blocks according to the image and volume resolutions. Key hyperparameters---including learning rates, batch sizes, selected block counts, and crop sizes across training stages---are detailed in Tab.~\ref{tab:hyperpara_supp}.

\begin{table}[t]
    \centering
    \caption{\textbf{Training details.} Hyperparameters across the three training stages. $M$ denotes the number of input images.}
    \label{tab:hyperpara_supp}
    {
    \setlength{\tabcolsep}{1pt}  
    \fontsize{6pt}{7pt}\selectfont
    \begin{tabular}{l c c c c c c}
        \toprule
        \multirow{2}{*}{\textbf{Stage}} & \multirow{2}{*}{Iterations} & Learning & Batch & Crop & \multicolumn{2}{c}{Top-$k$ Block Selection} \\
        \cmidrule(lr){6-7}
         &  & Rate & Size & Size & Vol $B^\text{i2v}, B^\text{v2v}$ & Img $B^\text{v2i}, B^\text{i2i}$ \\
        \midrule
        \textbf{Stage 1} & 16K & $4{\times}10^{-4} \rightarrow 1{\times}10^{-4}$ & 8 & 192 & N/A & N/A \\
        \textbf{Stage 2} & 12K & $1{\times}10^{-4} \rightarrow 4{\times}10^{-5}$ & 4 & 384 & 8 & $2M$ \\
        \textbf{Stage 3} & 3K & $4{\times}10^{-5} \rightarrow 1{\times}10^{-5}$ & 1 & 512 & 16 & $4M$ \\
        \bottomrule
    \end{tabular}
    }
\end{table}

Inverse-rendering fine-tuning follows the same stage-wise hyperparameter schedule, with the first two stages initialized from their NVS counterparts and trained for half of the listed iterations.

For novel view synthesis, we supervise rendered target views using appearance, foreground-mask, depth, and numerical normal losses from ground-truth synthetic 3D data. The appearance loss combines pixel-space MSE with LPIPS \cite{zhang2018unreasonable}, which we find important for preserving texture quality. All loss terms are evaluated by rendering the decoded volume representation into target views. To reduce the memory cost of LPIPS, we use deferred rendering \cite{zhang2022arf}, which caches full-image gradients and back-propagates through the MLP decoder and volume representation patch by patch. The novel-view synthesis objective is:
\begin{equation}
\begin{aligned}
\mathcal{L}_\text{NVS} =
{}&\mathcal{L}^\text{app}_\text{MSE} + 2\mathcal{L}^\text{app}_\text{LPIPS} + \mathcal{L}^\text{mask}_\text{MSE} \\
&+ 10\mathcal{L}^\text{depth}_\text{MSE} + 2\mathcal{L}^\text{normal}_\text{MSE}.
\end{aligned}
\end{equation}
We apply the numerical normal loss only during the final 500--1000 iterations to boost geometric fidelity, as it requires $3\times$ more samples during volume ray tracing. For inverse rendering, we replace the appearance loss with an albedo loss and further add MSE losses on the roughness and metallicity maps:
\begin{equation}
\begin{aligned}
\mathcal{L}_\text{IR} =
{}&\mathcal{L}^\text{albedo}_\text{MSE} + 2\mathcal{L}^\text{albedo}_\text{LPIPS} + \mathcal{L}^\text{rough}_\text{MSE} + \mathcal{L}^\text{metal}_\text{MSE} \\
&+ \mathcal{L}^\text{mask}_\text{MSE} + 10\mathcal{L}^\text{depth}_\text{MSE} + 2\mathcal{L}^\text{normal}_\text{MSE}.
\end{aligned}
\end{equation}
We compute all material losses in sRGB space, which empirically improves reconstruction quality compared to linear RGB space.

%% file: section/evaluation.tex
\section{Additional Evaluation Results}
\label{sec:app_evaluation}

\paragraph{Depth and normal comparisons} We further evaluate our normal and depth predictions on the DTC \cite{dong2025digital} dataset. Quantitative metrics are summarized in Tab. \ref{tab:inv_render_depth_normal}. Similar to recent feed-forward reconstruction methods \cite{li2025lirm}, LSRM demonstrates significantly greater robustness when handling highly specular surfaces compared to traditional optimization-based methods. Furthermore, our expanded context window yields noticeable improvements in fine geometric details over LIRM.

\begin{table}[t]
    \centering
    \begin{minipage}[c]{0.55\linewidth}
        \centering
       \setlength{\tabcolsep}{1pt}  
        \fontsize{6pt}{7pt}\selectfont
        {
        \begin{tabular}{l cc}
            \toprule
            \textbf{Method} & \textbf{Normal} ($\downarrow$) & \textbf{Depth} ($\downarrow$) \\
            \midrule
            NVDiffrecMc          & \cellcolor{yellow!15}0.06 & \cellcolor{yellow!15}0.02 \\
            InvRender            & \cellcolor{red!15}\textbf{0.03} & 0.22 \\
            LIRM                 & \cellcolor{orange!15}0.05 & \cellcolor{orange!15}0.017 \\
            \textbf{Ours (LSRM)} & \cellcolor{red!15}\textbf{0.03} & \cellcolor{red!15}\textbf{0.008} \\
            \bottomrule
        \end{tabular}
        }
    \end{minipage}\hfill
    \begin{minipage}[c]{0.4\linewidth}
        \caption{\textbf{Quantitative Geometry Evaluation.} Normal and depth reconstruction results on the DigitalTwinCatalogue \cite{dong2025digital} dataset.}
        \label{tab:inv_render_depth_normal}
    \end{minipage}
\end{table}

\begin{figure}[t]
\begin{minipage}[c]{0.55\linewidth}
    \centering
    \includegraphics[width=\columnwidth]{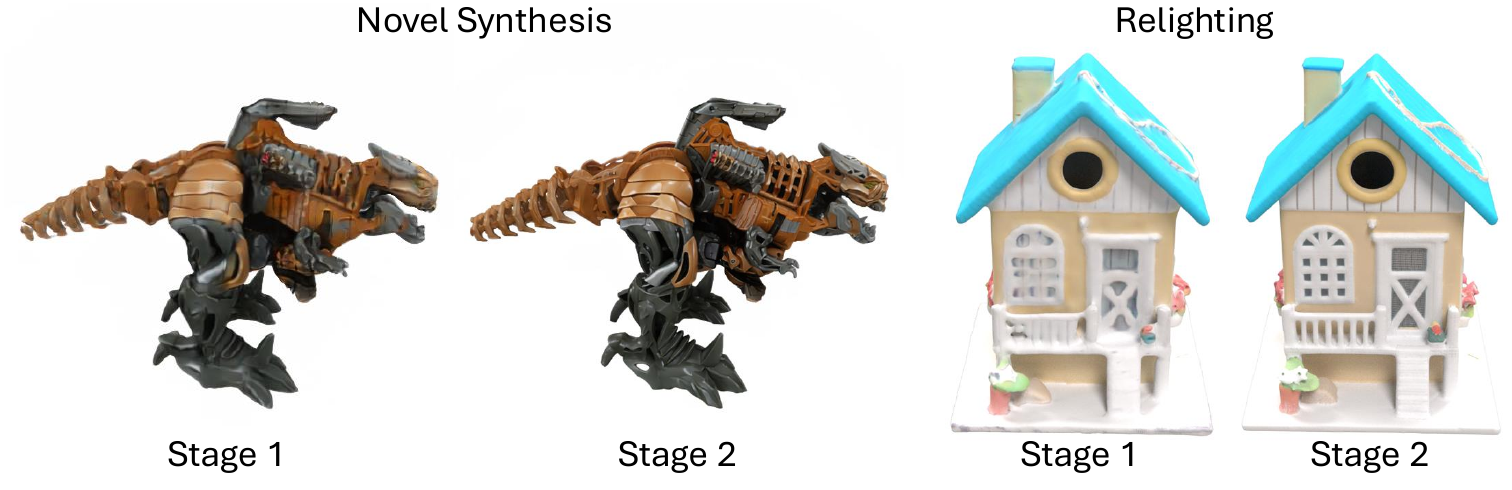}
\end{minipage}\hfill
\begin{minipage}[c]{0.4\linewidth}
    \caption{\textbf{Qualitative Geometry Evaluation.} LSRM is robust to geometric errors from Stage 1 reconstruction.}
    \label{fig:s1_geo_robustness}
\end{minipage}
\end{figure}

\paragraph{Robustness to Stage 1 geometric error:} The coarse geometric reconstruction not only provides the geometric cue for 3D-aware block routing, it also determines which tokens are activated in the Stage 2 sparse reconstruction transformer. We empirically find that our sparse volume initialization method is robust to geometric inaccuracies, as we intentionally select voxels that reside both on and \textit{near} the estimated surface, as shown in Eq.~\eqref{eq:surface_voxel_mask}. While this increases the token count, it is critical for high-fidelity reconstruction and allows the model to recover from coarse Stage 1 errors. Fig. \ref{fig:s1_geo_robustness} shows two examples where Stage 2 successfully refines complex structures where Stage 1 is imprecise, such as the thin fences of the birdhouse and the dinosaur's tail.

\paragraph{Comparison with the 3D Gaussian baseline} We compared LSRM against 3DGS on 10 randomly selected GSO models. As shown in Tab. \ref{tab:optimization}, LSRM using 8 to 16 views remains competitive with 3DGS using 64 views, while requiring orders of magnitude less time. Furthermore, LSRM significantly outperforms 3DGS when constrained to 16 or 32 views. This demonstrates that LSRM effectively narrows the quality gap between feed-forward and optimization-based methods.

\begin{table}[t]
    \centering
    \begin{minipage}[c]{0.5\linewidth}
        \centering
        \setlength{\tabcolsep}{1pt}  
        \fontsize{6pt}{7pt}\selectfont
        {
        \begin{tabular}{lcccccc}
            \toprule
            & \multicolumn{3}{c}{\textbf{3D Gaussian (3DGS)}} & & \multicolumn{2}{c}{\textbf{LSRM (Ours)}} \\
            \cmidrule{2-4} \cmidrule{6-7}
            \textbf{Metric} & 16 views & 32 views & 64 views & & 8 views & 16 views \\
            \midrule
            PSNR ($\uparrow$)  & 25.10 & 31.63 & \cellcolor{yellow!30}33.44 & & \cellcolor{orange!30}33.68 & \cellcolor{red!30}\textbf{34.17} \\
            SSIM ($\uparrow$)  & 0.944 & \cellcolor{orange!30}0.980 & \cellcolor{red!30}\textbf{0.985} & & 0.970 & \cellcolor{yellow!30}0.973 \\
            LPIPS ($\downarrow$) & 0.160 & 0.049 & \cellcolor{red!30}\textbf{0.022} & & \cellcolor{yellow!30}0.027 & \cellcolor{orange!30}0.025 \\
            \bottomrule
        \end{tabular}
        }
    \end{minipage}\hfill
    \begin{minipage}[c]{0.44\linewidth}
        \caption{\textbf{Quantitative NVS comparisons with optimization-based baselines.} Comparisons between 3DGS and LSRM on GSO show that LSRM narrows the quality gap between feed-forward and optimization-based methods.}
        \label{tab:optimization}
    \end{minipage}
\end{table}

%% file: main.bib
@String(ECCV  = {Eur. Conf. Comput. Vis.})

@String(TOG   = {ACM Trans. Graph.})

@String(ECCV  = {ECCV})

@String(TOG   = {ACM TOG})

@inproceedings{debevec2000acquiring,
  title={Acquiring the reflectance field of a human face},
  author={Debevec, Paul and Hawkins, Tim and Tchou, Chris and Duiker, Haarm-Pieter and Sarokin, Westley and Sagar, Mark},
  booktitle={Proceedings of the 27th annual conference on Computer graphics and interactive techniques},
  pages={145--156},
  year={2000}
}

@inproceedings{zhang2022iron,
  title={Iron: Inverse rendering by optimizing neural sdfs and materials from photometric images},
  author={Zhang, Kai and Luan, Fujun and Li, Zhengqi and Snavely, Noah},
  booktitle={Proceedings of the IEEE/CVF Conference on Computer Vision and Pattern Recognition},
  pages={5565--5574},
  year={2022}
}

@inproceedings{zhang2021physg,
  title={Physg: Inverse rendering with spherical gaussians for physics-based material editing and relighting},
  author={Zhang, Kai and Luan, Fujun and Wang, Qianqian and Bala, Kavita and Snavely, Noah},
  booktitle={Proceedings of the IEEE/CVF Conference on Computer Vision and Pattern Recognition},
  pages={5453--5462},
  year={2021}
}

@inproceedings{jin2023tensoir,
  title={Tensoir: Tensorial inverse rendering},
  author={Jin, Haian and Liu, Isabella and Xu, Peijia and Zhang, Xiaoshuai and Han, Songfang and Bi, Sai and Zhou, Xiaowei and Xu, Zexiang and Su, Hao},
  booktitle={Proceedings of the IEEE/CVF Conference on Computer Vision and Pattern Recognition},
  pages={165--174},
  year={2023}
}

@inproceedings{boss2021nerd,
  title={Nerd: Neural reflectance decomposition from image collections},
  author={Boss, Mark and Braun, Raphael and Jampani, Varun and Barron, Jonathan T and Liu, Ce and Lensch, Hendrik},
  booktitle={Proceedings of the IEEE/CVF International Conference on Computer Vision},
  pages={12684--12694},
  year={2021}
}

@article{boss2021neural,
  title={Neural-pil: Neural pre-integrated lighting for reflectance decomposition},
  author={Boss, Mark and Jampani, Varun and Braun, Raphael and Liu, Ce and Barron, Jonathan and Lensch, Hendrik},
  journal={Advances in Neural Information Processing Systems},
  volume={34},
  pages={10691--10704},
  year={2021}
}

@article{boss2022samurai,
  title={Samurai: Shape and material from unconstrained real-world arbitrary image collections},
  author={Boss, Mark and Engelhardt, Andreas and Kar, Abhishek and Li, Yuanzhen and Sun, Deqing and Barron, Jonathan and Lensch, Hendrik and Jampani, Varun},
  journal={Advances in Neural Information Processing Systems},
  volume={35},
  pages={26389--26403},
  year={2022}
}

@article{zhang2021nerfactor,
  title={Nerfactor: Neural factorization of shape and reflectance under an unknown illumination},
  author={Zhang, Xiuming and Srinivasan, Pratul P and Deng, Boyang and Debevec, Paul and Freeman, William T and Barron, Jonathan T},
  journal={ACM Transactions on Graphics (ToG)},
  volume={40},
  number={6},
  pages={1--18},
  year={2021},
  publisher={ACM New York, NY, USA}
}

@inproceedings{engelhardt2024shinobi,
  title={SHINOBI: Shape and Illumination using Neural Object Decomposition via BRDF Optimization In-the-wild},
  author={Engelhardt, Andreas and Raj, Amit and Boss, Mark and Zhang, Yunzhi and Kar, Abhishek and Li, Yuanzhen and Sun, Deqing and Brualla, Ricardo Martin and Barron, Jonathan T and Lensch, Hendrik and others},
  booktitle={Proceedings of the IEEE/CVF Conference on Computer Vision and Pattern Recognition},
  pages={19636--19646},
  year={2024}
}

@inproceedings{zhang2022modeling,
  title={Modeling indirect illumination for inverse rendering},
  author={Zhang, Yuanqing and Sun, Jiaming and He, Xingyi and Fu, Huan and Jia, Rongfei and Zhou, Xiaowei},
  booktitle={Proceedings of the IEEE/CVF Conference on Computer Vision and Pattern Recognition},
  pages={18643--18652},
  year={2022}
}

@inproceedings{munkberg2022extracting,
  title={Extracting triangular 3d models, materials, and lighting from images},
  author={Munkberg, Jacob and Hasselgren, Jon and Shen, Tianchang and Gao, Jun and Chen, Wenzheng and Evans, Alex and M{\"u}ller, Thomas and Fidler, Sanja},
  booktitle={Proceedings of the IEEE/CVF Conference on Computer Vision and Pattern Recognition},
  pages={8280--8290},
  year={2022}
}

@article{hasselgren2022shape,
  title={Shape, light, and material decomposition from images using monte carlo rendering and denoising},
  author={Hasselgren, Jon and Hofmann, Nikolai and Munkberg, Jacob},
  journal={Advances in Neural Information Processing Systems},
  volume={35},
  pages={22856--22869},
  year={2022}
}

@inproceedings{liang2024gs,
  title={Gs-ir: 3d gaussian splatting for inverse rendering},
  author={Liang, Zhihao and Zhang, Qi and Feng, Ying and Shan, Ying and Jia, Kui},
  booktitle={Proceedings of the IEEE/CVF Conference on Computer Vision and Pattern Recognition},
  pages={21644--21653},
  year={2024}
}

@inproceedings{jiang2024gaussianshader,
  title={Gaussianshader: 3d gaussian splatting with shading functions for reflective surfaces},
  author={Jiang, Yingwenqi and Tu, Jiadong and Liu, Yuan and Gao, Xifeng and Long, Xiaoxiao and Wang, Wenping and Ma, Yuexin},
  booktitle={Proceedings of the IEEE/CVF Conference on Computer Vision and Pattern Recognition},
  pages={5322--5332},
  year={2024}
}

@inproceedings{zhang2023nemf,
  title={Nemf: Inverse volume rendering with neural microflake field},
  author={Zhang, Youjia and Xu, Teng and Yu, Junqing and Ye, Yuteng and Jing, Yanqing and Wang, Junle and Yu, Jingyi and Yang, Wei},
  booktitle={Proceedings of the IEEE/CVF International Conference on Computer Vision},
  pages={22919--22929},
  year={2023}
}

@article{yang2023sire,
  title={Sire-ir: Inverse rendering for brdf reconstruction with shadow and illumination removal in high-illuminance scenes},
  author={Yang, Ziyi and Chen, Yanzhen and Gao, Xinyu and Yuan, Yazhen and Wu, Yu and Zhou, Xiaowei and Jin, Xiaogang},
  journal={arXiv preprint arXiv:2310.13030},
  year={2023}
}

@article{shi2023gir,
  title={Gir: 3d gaussian inverse rendering for relightable scene factorization},
  author={Shi, Yahao and Wu, Yanmin and Wu, Chenming and Liu, Xing and Zhao, Chen and Feng, Haocheng and Liu, Jingtuo and Zhang, Liangjun and Zhang, Jian and Zhou, Bin and others},
  journal={arXiv preprint arXiv:2312.05133},
  year={2023}
}

@article{careaga2023intrinsic,
  title={Intrinsic image decomposition via ordinal shading},
  author={Careaga, Chris and Aksoy, Ya{\u{g}}{\i}z},
  journal={ACM Transactions on Graphics},
  volume={43},
  number={1},
  pages={1--24},
  year={2023},
  publisher={ACM New York, NY, USA}
}

@inproceedings{meka2018lime,
  title={Lime: Live intrinsic material estimation},
  author={Meka, Abhimitra and Maximov, Maxim and Zollhoefer, Michael and Chatterjee, Avishek and Seidel, Hans-Peter and Richardt, Christian and Theobalt, Christian},
  booktitle={Proceedings of the IEEE conference on computer vision and pattern recognition},
  pages={6315--6324},
  year={2018}
}

@inproceedings{li2018cgintrinsics,
  title={Cgintrinsics: Better intrinsic image decomposition through physically-based rendering},
  author={Li, Zhengqi and Snavely, Noah},
  booktitle={Proceedings of the European conference on computer vision (ECCV)},
  pages={371--387},
  year={2018}
}

@inproceedings{li2018materials,
  title={Materials for masses: SVBRDF acquisition with a single mobile phone image},
  author={Li, Zhengqin and Sunkavalli, Kalyan and Chandraker, Manmohan},
  booktitle={Proceedings of the European conference on computer vision (ECCV)},
  pages={72--87},
  year={2018}
}

@article{li2018learning,
  title={Learning to reconstruct shape and spatially-varying reflectance from a single image},
  author={Li, Zhengqin and Xu, Zexiang and Ramamoorthi, Ravi and Sunkavalli, Kalyan and Chandraker, Manmohan},
  journal={ACM Transactions on Graphics (TOG)},
  volume={37},
  number={6},
  pages={1--11},
  year={2018},
  publisher={ACM New York, NY, USA}
}

@inproceedings{li2022physically,
  title={Physically-based editing of indoor scene lighting from a single image},
  author={Li, Zhengqin and Shi, Jia and Bi, Sai and Zhu, Rui and Sunkavalli, Kalyan and Ha{\v{s}}an, Milo{\v{s}} and Xu, Zexiang and Ramamoorthi, Ravi and Chandraker, Manmohan},
  booktitle={European Conference on Computer Vision},
  pages={555--572},
  year={2022},
  organization={Springer}
}

@inproceedings{sun2023neural,
  title={Neural-PBIR reconstruction of shape, material, and illumination},
  author={Sun, Cheng and Cai, Guangyan and Li, Zhengqin and Yan, Kai and Zhang, Cheng and Marshall, Carl and Huang, Jia-Bin and Zhao, Shuang and Dong, Zhao},
  booktitle={Proceedings of the IEEE/CVF International Conference on Computer Vision},
  pages={18046--18056},
  year={2023}
}

@article{li2017modeling,
  title={Modeling surface appearance from a single photograph using self-augmented convolutional neural networks},
  author={Li, Xiao and Dong, Yue and Peers, Pieter and Tong, Xin},
  journal={ACM Transactions on Graphics (ToG)},
  volume={36},
  number={4},
  pages={1--11},
  year={2017},
  publisher={ACM New York, NY, USA}
}

@article{deschaintre2018single,
  title={Single-image svbrdf capture with a rendering-aware deep network},
  author={Deschaintre, Valentin and Aittala, Miika and Durand, Fredo and Drettakis, George and Bousseau, Adrien},
  journal={ACM Transactions on Graphics (ToG)},
  volume={37},
  number={4},
  pages={1--15},
  year={2018},
  publisher={ACM New York, NY, USA}
}

@inproceedings{bi2020deep,
  title={Deep 3d capture: Geometry and reflectance from sparse multi-view images},
  author={Bi, Sai and Xu, Zexiang and Sunkavalli, Kalyan and Kriegman, David and Ramamoorthi, Ravi},
  booktitle={Proceedings of the IEEE/CVF conference on computer vision and pattern recognition},
  pages={5960--5969},
  year={2020}
}

@article{vaswani2017attention,
  title={Attention is all you need},
  author={Vaswani, A},
  journal={Advances in Neural Information Processing Systems},
  year={2017}
}

@article{hong2023lrm,
  title={Lrm: Large reconstruction model for single image to 3d},
  author={Hong, Yicong and Zhang, Kai and Gu, Jiuxiang and Bi, Sai and Zhou, Yang and Liu, Difan and Liu, Feng and Sunkavalli, Kalyan and Bui, Trung and Tan, Hao},
  journal={arXiv preprint arXiv:2311.04400},
  year={2023}
}

@article{wei2024meshlrm,
  title={Meshlrm: Large reconstruction model for high-quality mesh},
  author={Wei, Xinyue and Zhang, Kai and Bi, Sai and Tan, Hao and Luan, Fujun and Deschaintre, Valentin and Sunkavalli, Kalyan and Su, Hao and Xu, Zexiang},
  journal={arXiv preprint arXiv:2404.12385},
  year={2024}
}

@inproceedings{zhang2025gs,
  title={Gs-lrm: Large reconstruction model for 3d gaussian splatting},
  author={Zhang, Kai and Bi, Sai and Tan, Hao and Xiangli, Yuanbo and Zhao, Nanxuan and Sunkavalli, Kalyan and Xu, Zexiang},
  booktitle={European Conference on Computer Vision},
  pages={1--19},
  year={2025},
  organization={Springer}
}

@inproceedings{deitke2023objaverse,
  title={Objaverse: A universe of annotated 3d objects},
  author={Deitke, Matt and Schwenk, Dustin and Salvador, Jordi and Weihs, Luca and Michel, Oscar and VanderBilt, Eli and Schmidt, Ludwig and Ehsani, Kiana and Kembhavi, Aniruddha and Farhadi, Ali},
  booktitle={Proceedings of the IEEE/CVF Conference on Computer Vision and Pattern Recognition},
  pages={13142--13153},
  year={2023}
}

@article{zhang2024relitlrm,
  title={RelitLRM: Generative Relightable Radiance for Large Reconstruction Models},
  author={Zhang, Tianyuan and Kuang, Zhengfei and Jin, Haian and Xu, Zexiang and Bi, Sai and Tan, Hao and Zhang, He and Hu, Yiwei and Hasan, Milos and Freeman, William T and others},
  journal={arXiv preprint arXiv:2410.06231},
  year={2024}
}

@inproceedings{downs2022google,
  title={Google scanned objects: A high-quality dataset of 3d scanned household items},
  author={Downs, Laura and Francis, Anthony and Koenig, Nate and Kinman, Brandon and Hickman, Ryan and Reymann, Krista and McHugh, Thomas B and Vanhoucke, Vincent},
  booktitle={2022 International Conference on Robotics and Automation (ICRA)},
  pages={2553--2560},
  year={2022},
  organization={IEEE}
}

@article{kuang2024stanford,
  title={Stanford-ORB: a real-world 3d object inverse rendering benchmark},
  author={Kuang, Zhengfei and Zhang, Yunzhi and Yu, Hong-Xing and Agarwala, Samir and Wu, Elliott and Wu, Jiajun and others},
  journal={Advances in Neural Information Processing Systems},
  volume={36},
  year={2024}
}

@inproceedings{ummenhofer2024objects,
  title={Objects With Lighting: A Real-World Dataset for Evaluating Reconstruction and Rendering for Object Relighting},
  author={Ummenhofer, Benjamin and Agrawal, Sanskar and Sepulveda, Rene and Lao, Yixing and Zhang, Kai and Cheng, Tianhang and Richter, Stephan and Wang, Shenlong and Ros, German},
  booktitle={2024 International Conference on 3D Vision (3DV)},
  pages={137--147},
  year={2024},
  organization={IEEE}
}

@article{dao2023flashattention,
  title={Flashattention-2: Faster attention with better parallelism and work partitioning},
  author={Dao, Tri},
  journal={arXiv preprint arXiv:2307.08691},
  year={2023}
}

@article{yariv2021volume,
  title={Volume rendering of neural implicit surfaces},
  author={Yariv, Lior and Gu, Jiatao and Kasten, Yoni and Lipman, Yaron},
  journal={Advances in Neural Information Processing Systems},
  volume={34},
  pages={4805--4815},
  year={2021}
}

@inproceedings{zhang2018unreasonable,
  title={The unreasonable effectiveness of deep features as a perceptual metric},
  author={Zhang, Richard and Isola, Phillip and Efros, Alexei A and Shechtman, Eli and Wang, Oliver},
  booktitle={Proceedings of the IEEE conference on computer vision and pattern recognition},
  pages={586--595},
  year={2018}
}

@article{siddiqui2024meta,
  title={Meta 3d assetgen: Text-to-mesh generation with high-quality geometry, texture, and pbr materials},
  author={Siddiqui, Yawar and Monnier, Tom and Kokkinos, Filippos and Kariya, Mahendra and Kleiman, Yanir and Garreau, Emilien and Gafni, Oran and Neverova, Natalia and Vedaldi, Andrea and Shapovalov, Roman and others},
  journal={arXiv preprint arXiv:2407.02445},
  year={2024}
}

@inproceedings{zhang2022arf,
  title={Arf: Artistic radiance fields},
  author={Zhang, Kai and Kolkin, Nick and Bi, Sai and Luan, Fujun and Xu, Zexiang and Shechtman, Eli and Snavely, Noah},
  booktitle={European Conference on Computer Vision},
  pages={717--733},
  year={2022},
  organization={Springer}
}

@inproceedings{dong2025digital,
  title={Digital twin catalog: A large-scale photorealistic 3d object digital twin dataset},
  author={Dong, Zhao and Chen, Ka and Lv, Zhaoyang and Yu, Hong-Xing and Zhang, Yunzhi and Zhang, Cheng and Zhu, Yufeng and Tian, Stephen and Li, Zhengqin and Moffatt, Geordie and others},
  booktitle={Proceedings of the IEEE/CVF Conference on Computer Vision and Pattern Recognition},
  pages={753--763},
  year={2025}
}

@inproceedings{xiang2025structured,
  title={Structured 3d latents for scalable and versatile 3d generation},
  author={Xiang, Jianfeng and Lv, Zelong and Xu, Sicheng and Deng, Yu and Wang, Ruicheng and Zhang, Bowen and Chen, Dong and Tong, Xin and Yang, Jiaolong},
  booktitle={Proceedings of the IEEE/CVF conference on computer vision and pattern recognition},
  pages={21469--21480},
  year={2025}
}

@article{xiang2025native,
  title={Native and compact structured latents for 3d generation},
  author={Xiang, Jianfeng and Chen, Xiaoxue and Xu, Sicheng and Wang, Ruicheng and Lv, Zelong and Deng, Yu and Zhu, Hongyuan and Dong, Yue and Zhao, Hao and Yuan, Nicholas Jing and others},
  journal={arXiv preprint arXiv:2512.14692},
  year={2025}
}

@article{wu2025direct3d,
  title={Direct3d-s2: Gigascale 3d generation made easy with spatial sparse attention},
  author={Wu, Shuang and Lin, Youtian and Zhang, Feihu and Zeng, Yifei and Yang, Yikang and Bao, Yajie and Qian, Jiachen and Zhu, Siyu and Cao, Xun and Torr, Philip and others},
  journal={arXiv preprint arXiv:2505.17412},
  year={2025}
}

@article{gupta20233dgen,
  title={3dgen: Triplane latent diffusion for textured mesh generation},
  author={Gupta, Anchit and Xiong, Wenhan and Nie, Yixin and Jones, Ian and O{\u{g}}uz, Barlas},
  journal={arXiv preprint arXiv:2303.05371},
  year={2023}
}

@article{lan2025ln3diff++,
  title={LN3DIFF++: Scalable Latent Neural Fields Diffusion for Speedy 3D Generation},
  author={Lan, Yushi and Hong, Fangzhou and Zhou, Shangchen and Yang, Shuai and Meng, Xuyi and Chen, Yongwei and Lyu, Zhaoyang and Dai, Bo and Pan, Xingang and Loy, Chen Change},
  journal={IEEE Transactions on Pattern Analysis and Machine Intelligence},
  year={2025},
  publisher={IEEE}
}

@article{zhang20233dshape2vecset,
  title={3dshape2vecset: A 3d shape representation for neural fields and generative diffusion models},
  author={Zhang, Biao and Tang, Jiapeng and Niessner, Matthias and Wonka, Peter},
  journal={ACM Transactions On Graphics (TOG)},
  volume={42},
  number={4},
  pages={1--16},
  year={2023},
  publisher={ACM New York, NY, USA}
}

@article{hunyuan3d2025hunyuan3d,
  title={Hunyuan3d 2.1: From images to high-fidelity 3d assets with production-ready pbr material},
  author={Hunyuan3D, Team and Yang, Shuhui and Yang, Mingxin and Feng, Yifei and Huang, Xin and Zhang, Sheng and He, Zebin and Luo, Di and Liu, Haolin and Zhao, Yunfei and others},
  journal={arXiv preprint arXiv:2506.15442},
  year={2025}
}

@article{li2024craftsman3d,
  title={Craftsman3d: High-fidelity mesh generation with 3d native generation and interactive geometry refiner},
  author={Li, Weiyu and Liu, Jiarui and Yan, Hongyu and Chen, Rui and Liang, Yixun and Chen, Xuelin and Tan, Ping and Long, Xiaoxiao},
  journal={arXiv preprint arXiv:2405.14979},
  year={2024}
}

@article{li2025step1x,
  title={Step1x-3d: Towards high-fidelity and controllable generation of textured 3d assets},
  author={Li, Weiyu and Zhang, Xuanyang and Sun, Zheng and Qi, Di and Li, Hao and Cheng, Wei and Cai, Weiwei and Wu, Shihao and Liu, Jiarui and Wang, Zihao and others},
  journal={arXiv preprint arXiv:2505.07747},
  year={2025}
}

@article{zhang2024clay,
  title={Clay: A controllable large-scale generative model for creating high-quality 3d assets},
  author={Zhang, Longwen and Wang, Ziyu and Zhang, Qixuan and Qiu, Qiwei and Pang, Anqi and Jiang, Haoran and Yang, Wei and Xu, Lan and Yu, Jingyi},
  journal={ACM Transactions on Graphics (TOG)},
  volume={43},
  number={4},
  pages={1--20},
  year={2024},
  publisher={ACM New York, NY, USA}
}

@article{zhao2023michelangelo,
  title={Michelangelo: Conditional 3d shape generation based on shape-image-text aligned latent representation},
  author={Zhao, Zibo and Liu, Wen and Chen, Xin and Zeng, Xianfang and Wang, Rui and Cheng, Pei and Fu, Bin and Chen, Tao and Yu, Gang and Gao, Shenghua},
  journal={Advances in neural information processing systems},
  volume={36},
  pages={73969--73982},
  year={2023}
}

@inproceedings{yuan2025native,
  title={Native sparse attention: Hardware-aligned and natively trainable sparse attention},
  author={Yuan, Jingyang and Gao, Huazuo and Dai, Damai and Luo, Junyu and Zhao, Liang and Zhang, Zhengyan and Xie, Zhenda and Wei, Yuxing and Wang, Lean and Xiao, Zhiping and others},
  booktitle={Proceedings of the 63rd Annual Meeting of the Association for Computational Linguistics (Volume 1: Long Papers)},
  pages={23078--23097},
  year={2025}
}

@inproceedings{wang2024dust3r,
  title={Dust3r: Geometric 3d vision made easy},
  author={Wang, Shuzhe and Leroy, Vincent and Cabon, Yohann and Chidlovskii, Boris and Revaud, Jerome},
  booktitle={Proceedings of the IEEE/CVF conference on computer vision and pattern recognition},
  pages={20697--20709},
  year={2024}
}

@inproceedings{wang2025vggt,
  title={Vggt: Visual geometry grounded transformer},
  author={Wang, Jianyuan and Chen, Minghao and Karaev, Nikita and Vedaldi, Andrea and Rupprecht, Christian and Novotny, David},
  booktitle={Proceedings of the Computer Vision and Pattern Recognition Conference},
  pages={5294--5306},
  year={2025}
}

@inproceedings{leroy2024grounding,
  title={Grounding image matching in 3d with mast3r},
  author={Leroy, Vincent and Cabon, Yohann and Revaud, J{\'e}r{\^o}me},
  booktitle={European conference on computer vision},
  pages={71--91},
  year={2024},
  organization={Springer}
}

@article{lin2025depth,
  title={Depth anything 3: Recovering the visual space from any views},
  author={Lin, Haotong and Chen, Sili and Liew, Junhao and Chen, Donny Y and Li, Zhenyu and Shi, Guang and Feng, Jiashi and Kang, Bingyi},
  journal={arXiv preprint arXiv:2511.10647},
  year={2025}
}

@article{liang2024feed,
  title={Feed-forward bullet-time reconstruction of dynamic scenes from monocular videos},
  author={Liang, Hanxue and Ren, Jiawei and Mirzaei, Ashkan and Torralba, Antonio and Liu, Ziwei and Gilitschenski, Igor and Fidler, Sanja and Oztireli, Cengiz and Ling, Huan and Gojcic, Zan and others},
  journal={arXiv preprint arXiv:2412.03526},
  year={2024}
}

@article{ma20254d,
  title={4d-lrm: Large space-time reconstruction model from and to any view at any time},
  author={Ma, Ziqiao and Chen, Xuweiyi and Yu, Shoubin and Bi, Sai and Zhang, Kai and Ziwen, Chen and Xu, Sihan and Yang, Jianing and Xu, Zexiang and Sunkavalli, Kalyan and others},
  journal={arXiv preprint arXiv:2506.18890},
  year={2025}
}

@article{lin2025dgs,
  title={Dgs-lrm: Real-time deformable 3d gaussian reconstruction from monocular videos},
  author={Lin, Chieh Hubert and Lv, Zhaoyang and Wu, Songyin and Xu, Zhen and Nguyen-Phuoc, Thu and Tseng, Hung-Yu and Straub, Julian and Khan, Numair and Xiao, Lei and Yang, Ming-Hsuan and others},
  journal={arXiv preprint arXiv:2506.09997},
  year={2025}
}

@article{zhang2025efficiently,
  title={Efficiently reconstructing dynamic scenes one d4rt at a time},
  author={Zhang, Chuhan and Moing, Guillaume Le and Koppula, Skanda and Rocco, Ignacio and Momeni, Liliane and Xie, Junyu and Sun, Shuyang and Sukthankar, Rahul and Barral, Jo{\"e}lle K and Hadsell, Raia and others},
  journal={arXiv preprint arXiv:2512.08924},
  year={2025}
}

@article{xu20254dgt,
  title={4dgt: Learning a 4d gaussian transformer using real-world monocular videos},
  author={Xu, Zhen and Li, Zhengqin and Dong, Zhao and Zhou, Xiaowei and Newcombe, Richard and Lv, Zhaoyang},
  journal={arXiv preprint arXiv:2506.08015},
  year={2025}
}

@article{jin2024lvsm,
  title={Lvsm: A large view synthesis model with minimal 3d inductive bias},
  author={Jin, Haian and Jiang, Hanwen and Tan, Hao and Zhang, Kai and Bi, Sai and Zhang, Tianyuan and Luan, Fujun and Snavely, Noah and Xu, Zexiang},
  journal={arXiv preprint arXiv:2410.17242},
  year={2024}
}

@inproceedings{ziwen2025long,
  title={Long-lrm: Long-sequence large reconstruction model for wide-coverage gaussian splats},
  author={Ziwen, Chen and Tan, Hao and Zhang, Kai and Bi, Sai and Luan, Fujun and Hong, Yicong and Fuxin, Li and Xu, Zexiang},
  booktitle={Proceedings of the IEEE/CVF International Conference on Computer Vision},
  pages={4349--4359},
  year={2025}
}

@article{zhang2025test,
  title={Test-time training done right},
  author={Zhang, Tianyuan and Bi, Sai and Hong, Yicong and Zhang, Kai and Luan, Fujun and Yang, Songlin and Sunkavalli, Kalyan and Freeman, William T and Tan, Hao},
  journal={arXiv preprint arXiv:2505.23884},
  year={2025}
}

@inproceedings{li2025lirm,
  title={Lirm: Large inverse rendering model for progressive reconstruction of shape, materials and view-dependent radiance fields},
  author={Li, Zhengqin and Wang, Dilin and Chen, Ka and Lv, Zhaoyang and Nguyen-Phuoc, Thu and Lee, Milim and Huang, Jia-Bin and Xiao, Lei and Zhu, Yufeng and Marshall, Carl S and others},
  booktitle={Proceedings of the Computer Vision and Pattern Recognition Conference},
  pages={505--517},
  year={2025}
}

@inproceedings{zarzar2025twinner,
  title={Twinner: Shining Light on Digital Twins in a Few Snaps},
  author={Zarzar, Jesus and Monnier, Tom and Shapovalov, Roman and Vedaldi, Andrea and Novotny, David},
  booktitle={Proceedings of the Computer Vision and Pattern Recognition Conference},
  pages={5859--5869},
  year={2025}
}

@inproceedings{ainslie2023gqa,
  title={Gqa: Training generalized multi-query transformer models from multi-head checkpoints},
  author={Ainslie, Joshua and Lee-Thorp, James and De Jong, Michiel and Zemlyanskiy, Yury and Lebr{\'o}n, Federico and Sanghai, Sumit},
  booktitle={Proceedings of the 2023 Conference on Empirical Methods in Natural Language Processing},
  pages={4895--4901},
  year={2023}
}

@inproceedings{xu2024grm,
  title={Grm: Large gaussian reconstruction model for efficient 3d reconstruction and generation},
  author={Xu, Yinghao and Shi, Zifan and Yifan, Wang and Chen, Hansheng and Yang, Ceyuan and Peng, Sida and Shen, Yujun and Wetzstein, Gordon},
  booktitle={European Conference on Computer Vision},
  pages={1--20},
  year={2024},
  organization={Springer}
}

@inproceedings{chen2024lara,
  title={Lara: Efficient large-baseline radiance fields},
  author={Chen, Anpei and Xu, Haofei and Esposito, Stefano and Tang, Siyu and Geiger, Andreas},
  booktitle={European conference on computer vision},
  pages={338--355},
  year={2024},
  organization={Springer}
}

@article{goldman2009shape,
  title={Shape and spatially-varying brdfs from photometric stereo},
  author={Goldman, Dan B and Curless, Brian and Hertzmann, Aaron and Seitz, Steven M},
  journal={IEEE Transactions on Pattern Analysis and Machine Intelligence},
  volume={32},
  number={6},
  pages={1060--1071},
  year={2009},
  publisher={IEEE}
}

@inproceedings{zeng2024rgb,
author = {Zeng, Zheng and Deschaintre, Valentin and Georgiev, Iliyan and Hold-Geoffroy, Yannick and Hu, Yiwei and Luan, Fujun and Yan, Ling-Qi and Ha\v{s}an, Milo\v{s}},
title = {RGBX: Image decomposition and synthesis using material- and lighting-aware diffusion models},
year = {2024},
isbn = {9798400705250},
publisher = {Association for Computing Machinery},
address = {New York, NY, USA},
url = {https://doi.org/10.1145/3641519.3657445},
doi = {10.1145/3641519.3657445},
booktitle = {ACM SIGGRAPH 2024 Conference Papers},
articleno = {75},
numpages = {11},
location = {Denver, CO, USA},
series = {SIGGRAPH '24}
}

@article{wu2024direct3d,
  title={Direct3d: Scalable image-to-3d generation via 3d latent diffusion transformer},
  author={Wu, Shuang and Lin, Youtian and Zhang, Feihu and Zeng, Yifei and Xu, Jingxi and Torr, Philip and Cao, Xun and Yao, Yao},
  journal={Advances in Neural Information Processing Systems},
  volume={37},
  pages={121859--121881},
  year={2024}
}

@article{li2025triposg,
  title={Triposg: High-fidelity 3d shape synthesis using large-scale rectified flow models},
  author={Li, Yangguang and Zou, Zi-Xin and Liu, Zexiang and Wang, Dehu and Liang, Yuan and Yu, Zhipeng and Liu, Xingchao and Guo, Yuan-Chen and Liang, Ding and Ouyang, Wanli and others},
  journal={IEEE Transactions on Pattern Analysis and Machine Intelligence},
  year={2025},
  publisher={IEEE}
}

@inproceedings{chen20253dtopia,
  title={3dtopia-xl: Scaling high-quality 3d asset generation via primitive diffusion},
  author={Chen, Zhaoxi and Tang, Jiaxiang and Dong, Yuhao and Cao, Ziang and Hong, Fangzhou and Lan, Yushi and Wang, Tengfei and Xie, Haozhe and Wu, Tong and Saito, Shunsuke and others},
  booktitle={Proceedings of the Computer Vision and Pattern Recognition Conference},
  pages={26576--26586},
  year={2025}
}

@article{lan2025gaussiananything,
  title={Gaussiananything: Interactive point cloud latent diffusion for 3d generation},
  author={Lan, Yushi and Zhou, Shangchen and Lyu, Zhaoyang and Hong, Fangzhou and Yang, Shuai and Dai, Bo and Pan, Xingang and Loy, Chen Change},
  year={2025}
}

@article{vahdat2022lion,
  title={Lion: Latent point diffusion models for 3d shape generation},
  author={Vahdat, Arash and Williams, Francis and Gojcic, Zan and Litany, Or and Fidler, Sanja and Kreis, Karsten and others},
  journal={Advances in neural information processing systems},
  volume={35},
  pages={10021--10039},
  year={2022}
}

@article{yang2024atlas,
  title={Atlas gaussians diffusion for 3d generation},
  author={Yang, Haitao and Dong, Yuan and Jiang, Hanwen and Xu, Dejia and Pavlakos, Georgios and Huang, Qixing},
  journal={arXiv preprint arXiv:2408.13055},
  year={2024}
}

@article{li2025sparc3d,
  title={Sparc3d: Sparse representation and construction for high-resolution 3d shapes modeling},
  author={Li, Zhihao and Wang, Yufei and Zheng, Heliang and Luo, Yihao and Wen, Bihan},
  journal={arXiv preprint arXiv:2505.14521},
  year={2025}
}

@article{qiu2025gated,
  title={Gated attention for large language models: Non-linearity, sparsity, and attention-sink-free},
  author={Qiu, Zihan and Wang, Zekun and Zheng, Bo and Huang, Zeyu and Wen, Kaiyue and Yang, Songlin and Men, Rui and Yu, Le and Huang, Fei and Huang, Suozhi and others},
  journal={arXiv preprint arXiv:2505.06708},
  year={2025}
}

@misc{tillet2021triton,
  author       = {Tillet, Philippe},
  title        = {Introducing {Triton}: Open-source {GPU} programming for neural networks},
  year         = {2021},
  month        = {July},
  howpublished = {\url{https://openai.com/index/triton/}},
  note         = {OpenAI Blog}
}

@misc{lai2025nativesparseattention,
  author       = {Lai, Xunhao and Lu, Jianqiao},
  title        = {native-sparse-attention-triton: Efficient triton implementation of {Native Sparse Attention}},
  year         = {2025},
  publisher    = {GitHub},
  journal      = {GitHub repository},
  howpublished = {\url{https://github.com/XunhaoLai/native-sparse-attention-triton}}
}

@misc{wang2025nativesparseattentionpytorch,
  author       = {Wang, Phil},
  title        = {native-sparse-attention-pytorch: {Native Sparse Attention} in {PyTorch}},
  year         = {2025},
  publisher    = {GitHub},
  journal      = {GitHub repository},
  howpublished = {\url{https://github.com/lucidrains/native-sparse-attention-pytorch}}
}

@article{simeoni2025dinov3,
  title={Dinov3},
  author={Sim{\'e}oni, Oriane and Vo, Huy V and Seitzer, Maximilian and Baldassarre, Federico and Oquab, Maxime and Jose, Cijo and Khalidov, Vasil and Szafraniec, Marc and Yi, Seungeun and Ramamonjisoa, Micha{\"e}l and others},
  journal={arXiv preprint arXiv:2508.10104},
  year={2025}
}

@article{child2019generating,
  title={Generating long sequences with sparse transformers},
  author={Child, Rewon and Gray, Scott and Radford, Alec and Sutskever, Ilya},
  journal={arXiv preprint arXiv:1904.10509},
  year={2019}
}

@article{zaheer2020big,
  title={Big bird: Transformers for longer sequences},
  author={Zaheer, Manzil and Guruganesh, Guru and Dubey, Kumar Avinava and Ainslie, Joshua and Alberti, Chris and Ontanon, Santiago and Pham, Philip and Ravula, Anirudh and Wang, Qifan and Yang, Li and others},
  journal={Advances in neural information processing systems},
  volume={33},
  pages={17283--17297},
  year={2020}
}

@article{korthikanti2023reducing,
  title={Reducing activation recomputation in large transformer models},
  author={Korthikanti, Vijay Anand and Casper, Jared and Lym, Sangkug and McAfee, Lawrence and Andersch, Michael and Shoeybi, Mohammad and Catanzaro, Bryan},
  journal={Proceedings of Machine Learning and Systems},
  volume={5},
  pages={341--353},
  year={2023}
}

@inproceedings{loshchilov2018decoupled,
  title={Decoupled Weight Decay Regularization},
  author={Ilya Loshchilov and Frank Hutter},
  booktitle={International Conference on Learning Representations},
  year={2019},
  url={https://openreview.net/forum?id=Bkg6RiCqY7}
}

@inproceedings{henry2020query,
  title={Query-key normalization for transformers},
  author={Henry, Alex and Dachapally, Prudhvi Raj and Pawar, Shubham Shantaram and Chen, Yuxuan},
  booktitle={Findings of the Association for Computational Linguistics: EMNLP 2020},
  pages={4246--4253},
  year={2020}
}

@article{fang2024usp,
  title={Usp: A unified sequence parallelism approach for long context generative ai},
  author={Fang, Jiarui and Zhao, Shangchun},
  journal={arXiv preprint arXiv:2405.07719},
  year={2024}
}

@article{liu2023ring,
  title={Ring attention with blockwise transformers for near-infinite context},
  author={Liu, Hao and Zaharia, Matei and Abbeel, Pieter},
  journal={arXiv preprint arXiv:2310.01889},
  year={2023}
}

@article{jacobs2023deepspeed,
  title={Deepspeed ulysses: System optimizations for enabling training of extreme long sequence transformer models},
  author={Jacobs, Sam Ade and Tanaka, Masahiro and Zhang, Chengming and Zhang, Minjia and Song, Shuaiwen Leon and Rajbhandari, Samyam and He, Yuxiong},
  journal={arXiv preprint arXiv:2309.14509},
  year={2023}
}

@article{chen2025sam,
  title={Sam 3d: 3dfy anything in images},
  author={Chen, Xingyu and Chu, Fu-Jen and Gleize, Pierre and Liang, Kevin J and Sax, Alexander and Tang, Hao and Wang, Weiyao and Guo, Michelle and Hardin, Thibaut and Li, Xiang and others},
  journal={arXiv preprint arXiv:2511.16624},
  year={2025}
}
